%% file: Formatting-Instructions-LaTeX-2026.tex
\PassOptionsToPackage{dvipsnames}{xcolor}
\documentclass[letterpaper]{article} %
\usepackage{aaai2026}  %
\usepackage{times}  %
\usepackage{helvet}  %
\usepackage{courier}  %
\usepackage[hyphens]{url}  %
\usepackage{graphicx} %
\usepackage{natbib}  %
\usepackage{caption} %
\usepackage{algorithm}
\usepackage{algorithmic}

\usepackage{newfloat}
\usepackage{listings}
\DeclareCaptionStyle{ruled}{labelfont=normalfont,labelsep=colon,strut=off} %
\floatstyle{ruled}
\newfloat{listing}{tb}{lst}{}
\floatname{listing}{Listing}
\usepackage{blindtext}%

\usepackage{xstring}
\usepackage{bm}
\usepackage{multirow}
\usepackage{amssymb}%
\usepackage{pifont}%
\usepackage[flushleft]{threeparttable}
\usepackage{comment}
\usepackage{array}
\usepackage{colortbl}
\usepackage{cuted}
\usepackage{amsmath} 
\usepackage{cleveref}
\usepackage{booktabs} 
\usepackage{makecell}
\usepackage{subcaption}
\usepackage{url}

\title{Learning 3D Texture-Aware Representations for\\Parsing Diverse Human Clothing and Body Parts}
\author {
    Kiran Chhatre\textsuperscript{\rm 1,2},
    Christopher E. Peters\textsuperscript{\rm 1},
    Srikrishna Karanam\textsuperscript{\rm 2}
}
\affiliations {
    \textsuperscript{\rm 1}KTH Royal Institute of Technology \\
    \qquad \textsuperscript{\rm 2}Adobe Research\\
    \{chhatre, chpeters\}@kth.se, skaranam@adobe.com

}

\usepackage{bibentry}

\input{notation}

\nocopyright

\begin{document}

\maketitle

\input{sec/0_abstract}

\begin{links}
    \link{Extended version}{https://s-pectrum.github.io/}
\end{links}

\input{sec/1_intro}
\input{sec/2_relatedwork}

\input{sec/4_method}

\input{sec/5_implementation}

\input{sec/6_experiments}

\input{sec/7_conclusion}

\section{Acknowledgments}

We thank Michael J. Black for valuable feedback. We also thank Peter Kulits, Zitian Zhang, Zhening Huang, and Samuel Sartor for proofreading. This work was conducted during an internship at Adobe Research.

\bigskip
\newpage

\section*{Appendix}

\input{sec/X_suppl}

\bibliography{aaai2026}

\end{document}

%% file: notation.tex
\newcommand{\supmat}{Sup.~Mat.~}

\renewcommand{\paragraph}[1]{\noindent\textbf{#1}}

\newcommand{\formatcell}[1]{%
  \IfStrEq{#1}{--}{--}{%
    \StrSubstitute{#1}{,}{\quad}[\result]%
    \result%
  }%
}
\newcommand{\modelname}{Spectrum}

\newcommand{\clip}{\textsc{CLIP-Vision}~}  
\newcommand{\ic}{\textsc{im2context}~} 
\newcommand{\sd}{\textsc{SD}}
\definecolor{pastelblue}{RGB}{173, 216, 230}
\definecolor{softgreen}{RGB}{193, 225, 193}
\definecolor{peach}{RGB}{255, 218, 185}
\definecolor{lavender}{RGB}{230, 230, 250}

%% file: sec/0_abstract.tex
\begin{abstract}

Existing methods for human parsing into body parts and clothing often use fixed mask categories with broad labels that obscure fine-grained clothing types. Recent open-vocabulary segmentation approaches leverage pretrained text-to-image (T2I) diffusion model features for strong zero-shot transfer, but typically group entire humans into a single \textit{person} category, failing to distinguish diverse clothing or detailed body parts. To address this, we propose \modelname, a unified network for part-level pixel parsing (body parts and clothing) and instance-level grouping. While diffusion-based open-vocabulary models generalize well across tasks, their internal representations are not specialized for detailed human parsing. We observe that, unlike diffusion models with broad representations, image-driven 3D texture generators maintain faithful correspondence to input images, enabling stronger representations for parsing diverse clothing and body parts. \modelname~introduces a novel repurposing of an Image-to-Texture (I2Tx) diffusion model—obtained by fine-tuning a T2I model on 3D human texture maps—for improved alignment with body parts and clothing. From an input image, we extract human-part internal features via the I2Tx diffusion model and generate semantically valid masks aligned to diverse clothing categories through prompt-guided grounding. Once trained, \modelname~produces semantic segmentation maps for every visible body part and clothing category, ignoring standalone garments or irrelevant objects, for any number of humans in the scene. We conduct extensive cross-dataset experiments—separately assessing body parts, clothing parts, unseen clothing categories, and full-body masks—and demonstrate that \modelname~consistently outperforms baseline methods in prompt-based segmentation.

\end{abstract}

%% file: sec/1_intro.tex
\section{Introduction}
\label{sec:intro}

\begin{figure*}[t]
    \centering
    \includegraphics[width=\textwidth]{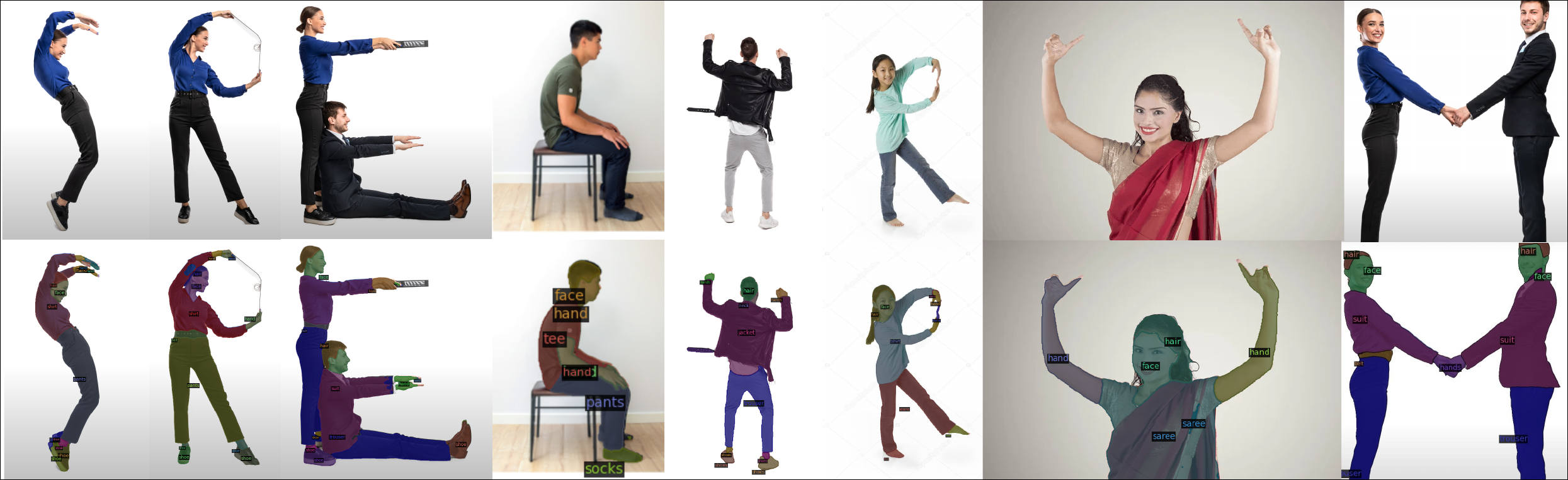}
    \caption{Human-alphabet parsing in the wild. Internet-video frames show people forming the letters of \modelname, while the model segments diverse clothing and body parts, handling unseen items such as \textcolor{blue}{\textit{sarees}}, across complex poses and styles.}
    \label{fig:teaser}
\end{figure*}

Human parsing~\cite{li2019selfcorrectionhumanparsing,fang2018weaklysemisupervisedhuman} decomposes a human image into distinct parts such as clothing items and body regions. This task provides rich descriptions for human-centric visual analysis~\cite{10.5555/3383741} and has become increasingly important in applications such as person re-identification~\cite{ye2021deep}, human-object interaction~\cite{peng2025hoi}, and portrait segmentation~\cite{wang2024eformer}. A detailed understanding of human body structure and the vast diversity in apparel is critical for advancing fields that analyze human appearance in various contexts, especially in complex scenes like those in~\cref{fig:teaser} (top), where the human alphabet forming our model name, \modelname, shows individuals wearing a range of formal, casual, and traditional attire.

Existing parsing methods typically rely on fixed, predefined categories. Although these are sufficient for human body parts, they struggle to handle the large variability and complexity of clothing in real world contexts, often producing broad or incorrect labels (e.g., labeling all clothing in the torso as `\textcolor{blue}{\textit{upper clothes}}'~\cite{khirodkar2024sapiensfoundationhumanvision}). Clothing fashion evolves rapidly, with new styles emerging while older ones become less common but still relevant, and accessories (e.g., \textcolor{blue}{\textit{hats}}, \textcolor{blue}{\textit{bags}}, \textcolor{blue}{\textit{belts}}, \textcolor{blue}{\textit{jewelry}}), as well as traditional attire (e.g., \textcolor{blue}{\textit{police uniforms}}, \textcolor{blue}{\textit{sarees}}), add further variation. Although recent high-quality annotated datasets~\cite{li2024cosmicman} capture this diversity, most current parsing architectures still work with a fixed set of labels, making it difficult to fully accommodate new or unseen clothing categories.

Emerging open-vocabulary approaches leverage diffusion models trained on large-scale text–image datasets, demonstrating strong compositional generalization and semantic control~\cite{Rombach_2022_CVPR}. However, they fail to distinguish body parts or fine-grained clothing, instead labeling the entire human as a single `\textcolor{blue}{\textit{person}}' mask in datasets such as COCO~\cite{coco}, because their broad representations are not specialized for detailed human parsing. Effective human parsing requires both part-level segmentation—capturing diverse clothing and distinct body parts—and instance-level grouping (e.g., distinguishing each individual's hands in multi-person scenes). Recent multimodal text and image driven 3D texture generation models show faithful correspondence to input modalities~\cite{tu2025smplgptexturedualview3dhuman}. We observe that these strong 3D texture-based features can significantly enhance segmentation of diverse human clothing and body parts. To address existing limitations in fine-grained human parsing, we leverage diffusion models fine-tuned on high-quality 3D human texture maps from the ATLAS dataset~\cite{liu2024texdreamer}, enabling accurate segmentation of diverse clothing categories and body parts.

In \modelname, we utilize these specialized internal representations that capture body parts and clothing features more effectively than standard diffusion model representations. We leverage the large-scale CosmicManHQ dataset~\cite{li2024cosmicman}, which contains detailed real-world images with text annotations describing human clothing and body parts, and focus exclusively on parsing on-body clothing, ignoring standalone garments or irrelevant items. Specifically, our model takes an image and a descriptive caption as input and outputs segmentation maps with semantic annotations for every visible body part and clothing category in scenes with any number of humans. To the best of our knowledge, our unified network is the first to offer this capability, which current human parsing and open-vocabulary segmentation methods lack. We use ground-truth masks for broad body-part categories and generic garment classes from CosmicManHQ to supervise the model, producing class-agnostic masks aligned with scene-specific prompts—enabling the parsing of both seen and unseen clothing categories. We validate our method through four evaluation setups: FPP (Full-Person Parsing, merging each person into one mask), BHP (Bare Human Parsing, separating each visible body part), CCP (COCO Category Parsing, focusing on human-relevant COCO accessory classes), and COP (Clothing-Only Parsing, isolating each clothing instance). Experiments on CosmicManHQ’s test set and cross-dataset scenarios demonstrate our model’s effectiveness in handling diverse clothing and body parts more effectively than existing human parsing and open-vocabulary segmentation approaches. As shown in~\cref{fig:teaser} (bottom), the model accurately segments formal (\textcolor{blue}{\textit{suit}}, \textcolor{blue}{\textit{trouser}}) and casual (\textcolor{blue}{\textit{shorts}}, \textcolor{blue}{\textit{jacket}}) attire in in-the-wild images, including unseen traditional clothing (\textcolor{blue}{\textit{sarees}}).

Our key contributions are: 
(1) We introduce \modelname, to the best of our knowledge the first human parsing approach that jointly addresses part-level pixel parsing for diverse on-body clothing, body parts, and instance-level part grouping via a unified network. 
(2) We repurpose an Image-to-Texture (I2Tx) diffusion model—fine-tuned from a T2I backbone on 3D human texture maps—to align internal features with diverse on-body clothing and body parts, enabling precise human parsing.
(3) Extensive cross-dataset experiments in four setups (FPP, BHP, CCP, COP) show that \modelname~consistently outperforms existing methods when parsing scenes with any number of humans.

%% file: sec/2_relatedwork.tex
\section{Related Work}
\label{sec:relatedworks}

\subsection{Human Parsing}

Early methods in human parsing focused on multiscale image features~\cite{chen2017deeplabsemanticimagesegmentation}, human pose~\cite{xia2016zoombetterclearerhuman}, or visual cues as weak supervision~\cite{dai2015boxsupexploitingboundingboxes}. Broadly, human parsing can be divided by the number of  human instances in the data and input modality: single-human parsing (e.g.,~\cite{dickens2025segmentationhumanmeshesmultiview}), multi-human instance-level parsing (e.g.,~\cite{liu2025schnetsammarriesclip}), and video-based parsing (e.g.,~\cite{gupta2023siamese}). Our scope is image-based single- and multi-human parsing. In single-human parsing, CDGNet~\cite{Liu_2022_CVPR} accumulates horizontal/vertical human-part distributions with attention for accurate relationship modeling, 
SCHP~\cite{li2019selfcorrectionhumanparsing} addresses annotation noise via self-correction, and SAPIENS~\cite{khirodkar2024sapiensfoundationhumanvision} exploits a ViT-based masked autoencoder pretrained on large-scale human images. In multi-human parsing, RP R-CNN~\cite{yang2020eccv} introduces a semantic-enhanced feature pyramid and parsing re-scoring network, while AIParsing~\cite{zhang2022aiparsing} takes a one-stage top-down approach for instance detection and part segmentation. Most previous methods rely on ground-truth image masks with fixed parsing labels, lacking dense annotations of human attributes and real-world complexity~\cite{gong2017lookpersonselfsupervisedstructuresensitive}, thus producing coarse results that cannot handle new categories to align masks categorically. We address this gap by pursuing robust fine-grained parsing for body parts and diverse clothing, incorporating instance-level part grouping in a unified network.

\subsection{Open-Vocabulary Segmentation}

\paragraph{Background.} Open-domain tasks have gained increasing attention for their ability to handle unseen categories during inference~\cite{10.1109/TPAMI.2024.3413013}. These include open-set recognition~\cite{Geng_2021}, open-world recognition~\cite{kim2021learningopenworldobjectproposals}, and open-vocabulary learning~\cite{wu2024openvocabularylearningsurvey}. Large vision–language models (VLMs) trained with contrastive image and text objectives provide strong zero shot transfer~\cite{jia2021scalingvisualvisionlanguagerepresentation}, and T2I models further bind region–word correspondences~\cite{ho2020denoisingdiffusionprobabilisticmodels}. Below, we review open-vocabulary semantic and instance segmentation methods relevant to our work.

\paragraph{Semantic segmentation.} Recent open-vocabulary semantic segmentation methods adapt pretrained VLMs or diffusion models to segment novel classes. ODISE~\cite{xu2023odise} uses frozen T2I diffusion features and an implicit captioner, while MaskCLIP~\cite{ding2023maskclip} modifies CLIP’s final attention layer for stronger local semantic consistency. IFSeg~\cite{yun2023ifseg} processes target categories given only label sets, and GD~\cite{li2023grounded} aligns T2I model’s visual and textual embeddings via a grounding module. OVSeg~\cite{liang2023open} fine-tunes CLIP with mask–category pairs to overcome domain gaps, and OVAM~\cite{Marcos-Manchon_2024_CVPR} generates word-specific attention maps in T2I models without additional training. SED~\cite{xie2024sed} applies a hierarchical encoder–decoder for cost-map generation, while SEEM~\cite{zou2023segment} employs prompt-based interactive segmentation. OpenSeg~\cite{10.1007/978-3-031-20059-5_31} randomly drops words during training to prevent overfitting. Broadly, some T2I-based methods synthesize labeled images~\cite{li2022bigdatasetgansynthesizingimagenetpixelwise} or perform image conditioning for mask generation~\cite{ji2023ddpdiffusionmodeldense}.

\paragraph{Instance segmentation.} GGN~\cite{wang2022ggn} uses pseudo supervision from learned pixel-level affinities,  OLN~\cite{kim2021oln} applies classifier-free object localization, and OV-SAM~\cite{10.1007/978-3-031-72775-7_24} merges CLIP and SAM for open-vocabulary recognition. Despite the successes of open-vocabulary methods, these approaches rarely address fine-grained human body parts or complex clothing categories, often collapsing people into a single \textcolor{blue}{\textit{person}} mask. We bridge this gap by adapting open-vocabulary segmentation to detailed body-part and clothing parsing for arbitrary humans. To this end, we adopt a fine-tuned T2I model on human textures, yielding specialized representations that enable accurate segmentation of detailed body parts and diverse clothing.

%% file: sec/4_method.tex
\section{Method}
\label{sec:method}

\begin{figure}[t]
  \centering
  \includegraphics[width=0.6\linewidth]{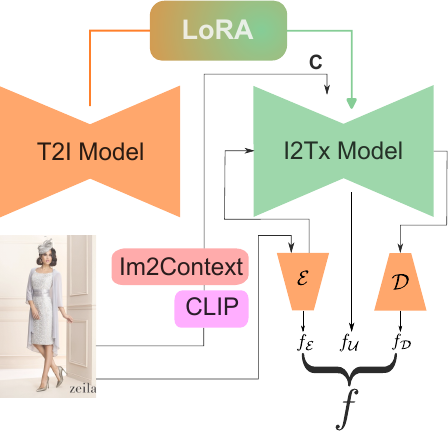}
  \caption{I2Tx feature extraction. Base Stable Diffusion (\sd) weights are merged with 3D-texture LoRA matrices, and a context embedding $\mathbf{C}$ from \ic\ and \clip\ drives a single forward pass in the I2Tx model. Concatenating encoder, denoiser, and decoder representations yields the texture-aligned feature $f$ for clothing and body-part parsing.}

  \label{fig:i2tx}
\end{figure}

\paragraph{Overview.} We focus on prompt-based human parsing for diverse clothing categories and individual body parts. During training, ground truth masks for 17 generic base categories and dense captions are provided; body part masks are detailed, whereas clothing masks are weakly annotated because the base categories (e.g., \textcolor{blue}{\textit{Special Clothings}}, \textcolor{blue}{\textit{One-piece Outfits}}) are broad and require grounding to learn accurate labels. At test time, clothing may include both seen and unseen classes, and we ensure that the test subset explicitly includes unseen categories for separate evaluation.

\subsection{Repurposing Image-to-Texture Models}

\paragraph{Preliminaries.} We leverage strong 3D texture-based features with faithful correspondence to input images to enhance segmentation of diverse clothing and body parts. Consequently, we repurpose diffusion features from TexDreamer~\cite{liu2024texdreamer}, a dual-conditioned (text/image) 3D texture generator, for 2D human parsing. TexDreamer builds on Stable Diffusion (\sd)~\cite{Rombach_2022_CVPR} with LoRA, trained on 3D human texture maps paired with images and captions from the ATLAS dataset. It conditions on image features from the \ic~encoder, whose LoRA weights and pretrained encoder we reuse in our method.

\paragraph{I2Tx feature extraction.} As shown in~\cref{fig:i2tx}, our feature-extraction pipeline uses the frozen I2Tx architecture in a single forward pass ($t=0$), omitting iterative denoising. We first merge the base \sd~weights $W_{\phi_{\text{SD}}}\!\in\!\mathbb{R}^{d\times k}$ with TexDreamer’s LoRA matrices $B\!\in\!\mathbb{R}^{d\times r}$ and $A\!\in\!\mathbb{R}^{r\times k}$ (with $r\!\ll\!\min(d,k)$), so the adapted projection becomes $W_{\phi_{\text{SD}}}+BA$. An input image $x$ is then encoded by the \clip~encoder $\phi_{\text{CV}}$ and the \ic~encoder $\phi_{\text{I2C}}$ to generate the context embedding $\mathbf{C}$ and the \textsc{CLS} token.
\begin{equation}
\textbf{C} = \phi_{\text{I2C}} \left\{ \phi_{\text{CV}}(x) \right\}
\label{eq:context}
\end{equation}
The encoder $\phi_{\mathcal{E}}$ produces features $f_{\mathcal{E}}$ and a latent code $x_e$. A noisy latent is sampled with noise schedule $\bar{\alpha}_{t}=\prod_{k=1}^{t}\alpha_k$.
\begin{equation}
\begin{aligned}
f_{\mathcal{E}}, x_{e} &= \phi_{\mathcal{E}}(x),   \\
x_{t} &\triangleq \sqrt{\overline{\alpha}_{t}}x_{e} + \sqrt{1-\overline{\alpha}_{t}}\epsilon, \quad \epsilon \sim \mathcal{N}(0,\textbf{I}),
\end{aligned}
\end{equation}
The denoiser $\phi_{\text{SD}}$ is conditioned on $(x_{t},\mathbf{C},\text{CLS})$ to yield $f_{U}$, while the decoder $\phi_{\mathcal{D}}$ processes $x_{t}$ to obtain $f_{\mathcal{D}}$. Concatenating the encoder, denoiser, and decoder streams yields a texture-aligned representation \(f\) that drives our segmentation head to parse on-body clothing and body parts.
\begin{equation}
\begin{aligned}
f_{U} &= \phi_{\text{SD}}(x_{t}, \textbf{C}, \text{CLS}) \\
f_{\mathcal{D}} &= \phi_{\mathcal{D}}(x_{t}) \\
f &= f_{\mathcal{E}} \parallel f_{U} \parallel f_{\mathcal{D}}
\end{aligned}
\end{equation}

\subsection{Semantically Grounded Parsing}

\begin{figure}[t]
  \centering
  \includegraphics[width=1\linewidth]{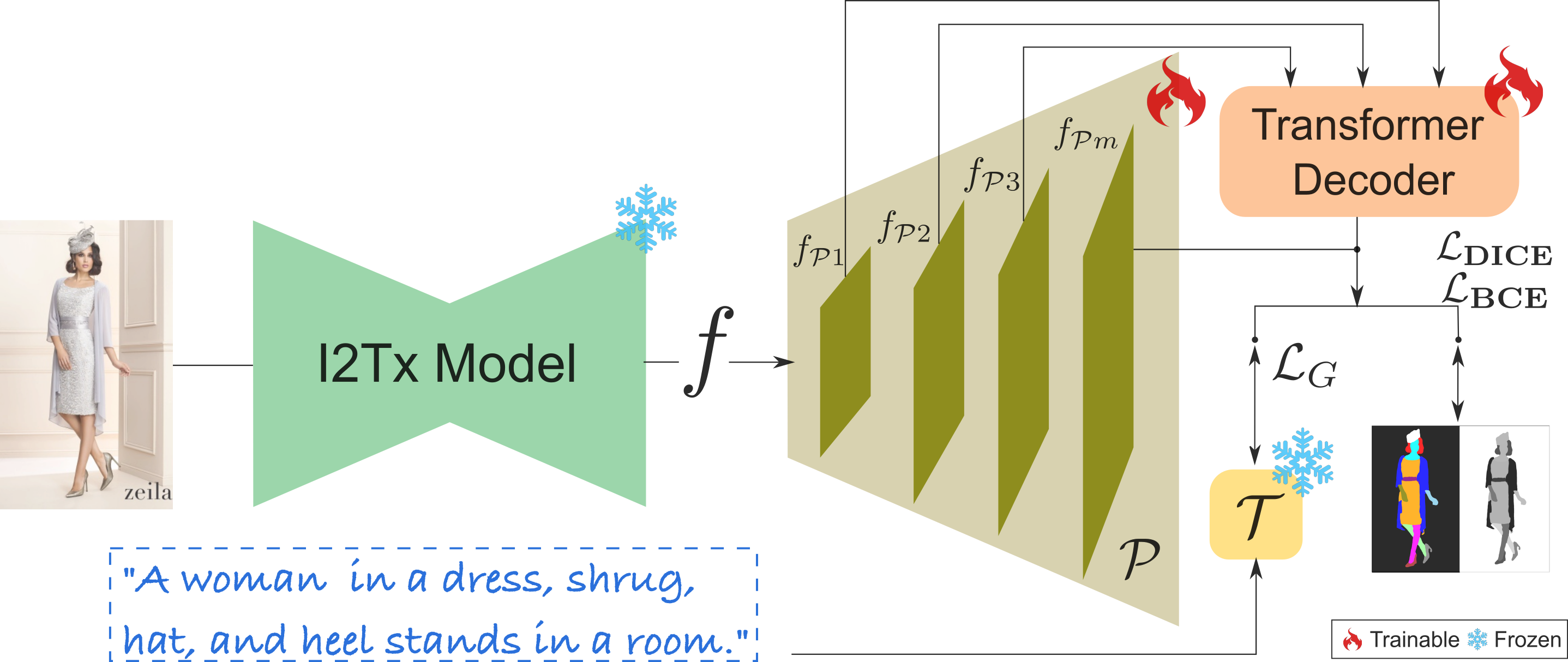}
  \caption{Training. Frozen I2Tx features from image $x$ feed the pixel decoder $\mathcal{P}$ and transformer, producing $N$ class-agnostic masks $m$ optimized with $\mathcal{L}_{\mathrm{BCE}}$ and $\mathcal{L}_{\mathrm{DICE}}$. Key prompt phrases, embedded by frozen $\mathcal{T}$, are contrastively aligned to $m$ via $\mathcal{L}_{\mathrm{G}}$.}

  \label{fig:arch_feat}
  \vspace{-10pt}
\end{figure}

\paragraph{Parsing-head architecture.}
As shown in~\cref{fig:arch_feat}, a pixel decoder $\mathcal{P}$ (implemented as a feature-pyramid network~\cite{cheng2021mask2former}) converts the texture representation $f$ into multi-scale, high-resolution per-pixel embeddings. Its final mask feature $f_{\mathcal{P}m}$ together with the pyramid outputs $f_{\mathcal{P}1}, f_{\mathcal{P}2}, f_{\mathcal{P}3}$ feeds a transformer decoder that predicts $N$ class-agnostic binary masks $\{m_i\}_{i=1}^{N}$.
For each mask $m_i$, we obtain its embedding via masked average pooling, $z_i=\langle f, m_i\rangle$. We also extract auxiliary mask embeddings and pyramid features in the same manner. Following \citet{cheng2021perpixelclassificationneedsemantic}, Hungarian matching assigns predictions to ground-truth masks. Class logits are optimized with binary cross-entropy loss $\mathcal{L}_{\mathrm{BCE}}$, and mask shape with Dice loss $\mathcal{L}_{\mathrm{DICE}}$.

\paragraph{Prompt grounding.}
We train on CosmicMan-HQ, which provides BLIP-generated dense captions containing detailed hair and clothing attributes. For each image–caption pair $\{x^{(m)}, s^{(m)}\}_{m=1}^{B}$ in a batch of size $B$, we extract $K_{\text{phrase}}$ key phrases (relevant nouns and adjectives) from $s^{(m)}$, yielding $\{p_k\}_{k=1}^{K_{\text{phrase}}}$. Each phrase $p_k$ is formatted into a prompt (e.g., “\textcolor{blue}{\textit{a photo of a tan purse}}” or “\textcolor{blue}{\textit{person hair length is above shoulders}}”) and embedded with \textsc{OpenCLIP} ($\mathcal{T}$). Following~\cite{xu2023odise}, we apply a contrastive loss that aligns phrase embeddings $\mathcal{T}(p_k)$ with their corresponding mask embeddings $z_i$. The grounding loss $\mathcal{L}_{G}$ over the batch is then computed as:
\begin{equation}
\begin{split}
    g\!\left(x^{(m)},\, s^{(m)}\right)
 = \frac{1}{K} \sum_{k=1}^{K} \sum_{i=1}^{N} \mathbf{p}_{ik} \cdot \langle z_i, \mathcal{T}(p_k) \rangle, \\
    \mathcal{L}_G = - \frac{1}{B} \sum_{m=1}^{B} \log \frac{e^{2 \cdot g_{mm}/\tau}}{\sum_{n=1}^{B} e^{g_{mn}/\tau} \cdot \sum_{n=1}^{B} e^{g_{nm}/\tau}}.
\end{split}
\label{eq:gloss}
\end{equation}

Here, $\mathrm{\textbf{p}}_{ik}$ is the softmax probability that mask embedding \(z_i\) matches base phrase \(p_k\), with \(N\) denoting the number of class-agnostic binary masks. The similarity between mask and text embeddings is given by \(\langle z_i, \mathcal{T}(p_k) \rangle\). We use \(g_{mm}\) as a shorthand for \(g(x^{(m)}, s^{(m)})\), the similarity score for the matched image-caption pair, while \(g_{mn}\) and \(g_{nm}\) denote cross-similarity scores within the batch—\(g_{mn}\) being the similarity between the \(m\)-th image and other captions \(s^n\), and \(g_{nm}\) the similarity between the \(m\)-th caption and other images \(x^n\). \(\tau\) is a learnable temperature scaling the similarity scores. The total loss \(\mathcal{L}_{\text{total}}\) is computed using weights \(\lambda_{\text{BCE}}\), \(\lambda_{\text{DICE}}\), and \(\lambda_{\text{G}}\) for \(\mathcal{L}_{\text{BCE}}\), \(\mathcal{L}_{\text{DICE}}\), and \(\mathcal{L}_{\text{G}}\), respectively.

\begin{equation}
\begin{aligned}
\mathcal{L}_{\text{BCE}} &= \text{BCE}(\mathbf{p},\mathbf{y}), \quad
\mathcal{L}_{\text{DICE}} = 1 - \frac{2\,\mathbf{y}\cdot\mathbf{p}+1}{\|\mathbf{y}\|+\|\mathbf{p}\|+1}, \\[1mm]
\mathcal{L}_{\text{total}} &= 
\lambda_{\text{BCE}}\mathcal{L}_{\text{BCE}}
+ \lambda_{\text{DICE}}\mathcal{L}_{\text{DICE}}
+ \lambda_{\text{G}}\mathcal{L}_{\text{G}}.
\end{aligned}
\end{equation}

\noindent Here, \(y_{i}\) is the ground-truth mask. In \(\mathcal{L}_{\text{DICE}}\), the numerator is the pixel-wise dot product (\(\sum_{i=1}^{N} y_i\, \mathbf{p}_{ik}\)), and the denominator sums the pixel values of \(\mathbf{y}\) and \(\mathbf{p}_{ik}^{b}\).

%% file: sec/5_implementation.tex
\section{Implementation Details}
\label{sec:implementation}

\begin{table*}[t]
    \centering
    {\scriptsize
    \resizebox{\textwidth}{!}{%
        \begin{tabular}{l|cccccc|cccccc}
            \multirow{3}{*}{\textbf{Method}} & \multicolumn{6}{c|}{\textbf{CosmicManHQ}} & \multicolumn{6}{c}{\textbf{Cross-dataset: GranD-f}} \\
            & \multicolumn{3}{c}{\textbf{COP}} & \multicolumn{3}{c|}{\textbf{BHP}} & \multicolumn{3}{c}{\textbf{COP}} & \multicolumn{3}{c}{\textbf{BHP}} \\
            & \textbf{mIoU} & \textbf{mAcc} & \textbf{mAP\textsuperscript{SS}} & \textbf{mIoU} & \textbf{mAcc} & \textbf{mAP\textsuperscript{SS}} &
              \textbf{mIoU} & \textbf{mAcc} & \textbf{mAP\textsuperscript{SS}} & \textbf{mIoU} & \textbf{mAcc} & \textbf{mAP\textsuperscript{SS}} \\
            \Xhline{.8pt}
            PPP-SCHP~\cite{mottaghi_cvpr14} & --- & --- & --- & 45.2 & 49.7 & 34.0 & --- & --- & --- & 20.4 & 22.4 & 15.3 \\
            ATR-SCHP~\cite{Liang_2015} & 44.3 & 48.7 & 34.0 & 50.2 & 55.4 & 37.6 & 19.5 & 21.4 & 14.7 & 22.8 & 25.1 & 17.2 \\
            LIP-SCHP~\cite{gong2017lookpersonselfsupervisedstructuresensitive} & 42.0 & 45.4 & 31.1 & 46.6 & 50.8 & 34.7 & 18.5 & 20.3 & 13.9 & 21.6 & 23.7 & 16.2 \\
            CIHP-PGN~\cite{gong2018instancelevelhumanparsinggrouping} & 43.1 & 47.1 & 30.6 & 48.4 & 53.1 & 37.1 & 19.0 & 20.8 & 13.3 & 22.2 & 23.4 & 16.0 \\
            Sapiens-1B~\cite{khirodkar2024sapiensfoundationhumanvision} &
            \cellcolor{gray!20}52.4 & \cellcolor{gray!20}57.1 & \cellcolor{gray!20}40.3 &
            \cellcolor{gray!20}58.5 & \cellcolor{gray!20}62.2 & \cellcolor{gray!20}42.1 &
            \cellcolor{gray!20}25.1 & \cellcolor{gray!20}27.3 & \cellcolor{gray!20}18.4 &
            \cellcolor{gray!20}27.5 & \cellcolor{gray!20}30.5 & \cellcolor{gray!20}20.5 \\
            M2F-C\textsubscript{close}~\cite{cheng2021mask2former} & 49.7 & 54.4 & 33.6 & 54.1 & 57.2 & 34.1 & 24.8 & 26.9 & 17.6 & 25.0 & 27.0 & 18.0 \\
            ODISE-C\textsubscript{open}~\cite{xu2023odise} &
            \cellcolor{gray!20}65.7 & \cellcolor{gray!20}72.2 & \cellcolor{gray!20}49.3 &
            \cellcolor{gray!20}73.4 & \cellcolor{gray!20}77.0 & \cellcolor{gray!20}53.7 &
            \cellcolor{gray!20}21.5 & \cellcolor{gray!20}27.2 & \cellcolor{gray!20}12.2 &
            \cellcolor{gray!20}23.5 & \cellcolor{gray!20}29.2 & \cellcolor{gray!20}14.2 \\
            Ours &
            \cellcolor{pastelblue!60}\textbf{77.5} & \cellcolor{pastelblue!60}\textbf{79.4} & \cellcolor{pastelblue!60}\textbf{58.2} &
            \cellcolor{pastelblue!60}\textbf{86.3} & \cellcolor{pastelblue!60}\textbf{88.2} & \cellcolor{pastelblue!60}\textbf{65.1} &
            \cellcolor{pastelblue!60}\textbf{27.1} & \cellcolor{pastelblue!60}\textbf{29.8} & \cellcolor{pastelblue!60}\textbf{21.0} &
            \cellcolor{pastelblue!60}\textbf{30.2} & \cellcolor{pastelblue!60}\textbf{33.1} & \cellcolor{pastelblue!60}\textbf{22.9} \\
        \end{tabular}%
    }}
    \caption{Human parsing on CosmicManHQ (test) and GranD-f (cross) under COP and BHP. M2F-C\textsubscript{close} and ODISE-C\textsubscript{open} are retrained on our CosmicManHQ split. Best scores are highlighted in \colorbox{pastelblue!60}{\textbf{teal}}, and second-best scores are shown in \colorbox{gray!20}{gray}.}
    \label{tab:close_set}
    \vspace{-10pt}
\end{table*}

\paragraph{I2Tx model.} We use \textsc{CLIP-ViT-H/14} (trained on the 224×224 LAION-2B English subset of LAION-5B) as the base \clip model. TexDreamer trains an \ic model (linear projection + 3-layer Transformer decoder; visual/text dims: 1280/1024) and fine-tunes \clip (scaling factor \(\alpha=16\), rank \(r=16\)) and \sd~(\(\alpha=128\), \(r=128\)) with LoRA, all on the ATLAS dataset~\cite{liu2024texdreamer} (50K 1024×1024 3D human textures in SMPL UV space). We merge the LoRA adapters into \clip and the texture-finetuned SD UNet via Diffusers, using a diffusion time step of 0, a 256-dimensional latent, and seed 777. Captions are processed by extracting \(K_{\text{phrase}}=9\) nouns/adjectives with NLTK, and text embeddings come from \textsc{ViT-H/14 OpenCLIP}.

\paragraph{Semantic parsing.} I2Tx features are upsampled via \textsc{MSDeformAttn}~\cite{zhu2021deformabledetrdeformabletransformers}. The transformer decoder (8 heads, 9 layers) predicts $N=100$ class-agnostic binary masks and dynamically assigns unique colours to each mask for body parts and clothing.

\paragraph{Datasets.} We train on 100K images randomly drawn from 27 parquet files (514K total) in LAION2B-en and LAION1B-nolang (CosmicMan-HQ subset of LAION-5B), yielding 6M images overall, obtained via img2dataset tool. Seventeen base categories are listed in~\cref{tab:per_class_iou}, and GT masks are resized to the original image size during training.

\paragraph{Training details.} Images are augmented with horizontal/vertical flips and resized to 1024; test images keep aspect ratio with a 1024 short side. Loss weights are \(\lambda_{\text{BCE}}{=}2.0\), \(\lambda_{\text{DICE}}{=}5.0\), \(\lambda_{\text{G}}{=}1.0\). Following~\cite{cheng2021mask2former}, mask loss is computed on 112\(\times\)112 (12,544) random points from matched masks. We use AdamW with a 1e-4 learning rate and 0.05 weight decay. The model has 2.005B parameters (29.55M trainable, 0.86\%) and trains for 12.2 days (370K iters, batch size 8) on eight A100 GPUs.

\begin{figure}[b]
  \centering
   \includegraphics[width=\linewidth]{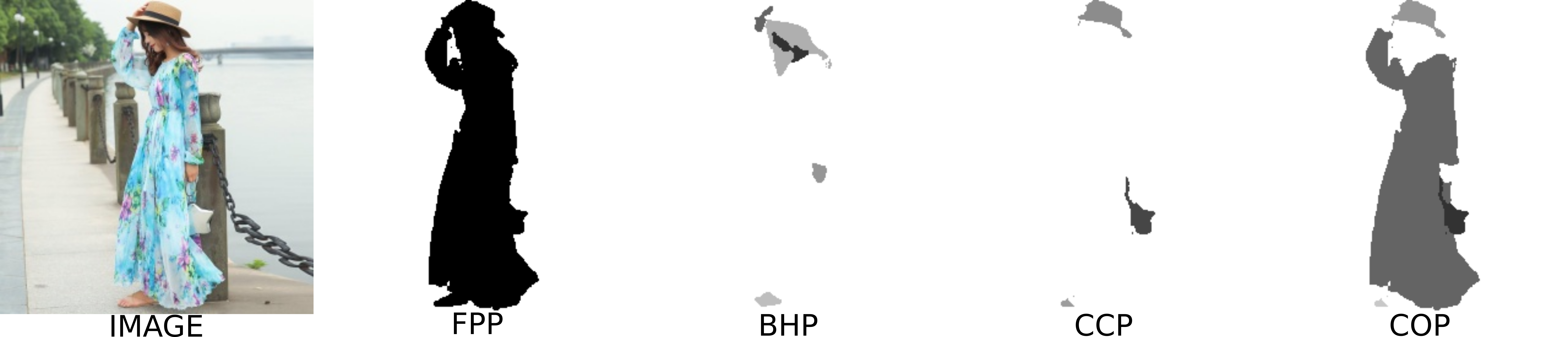}
   \caption{FPP, BHP, CCP, and COP masks.}
   
   \label{fig:eval_setting}
\end{figure}

\begin{figure}[t]
    \centering
    \includegraphics[width=\linewidth]{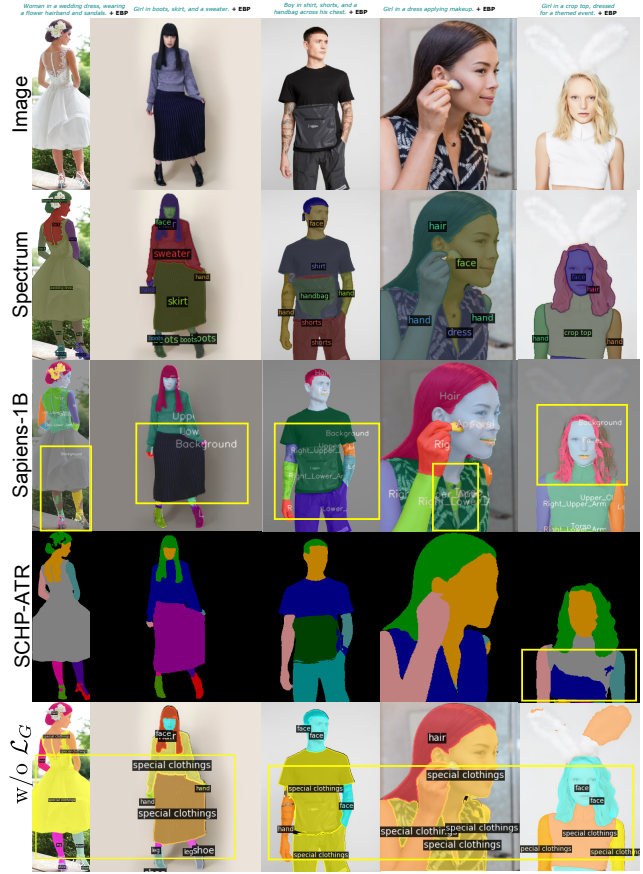}
    \caption{Standard parsers (R2–4); w/o \(\mathcal{L}_G\) ablation (\mbox{R5}).}

    \label{fig:cosmic_results}
    \vspace{-10pt}
\end{figure}

\begin{figure*}[t]
    \centering
    \includegraphics[width=\textwidth,keepaspectratio]{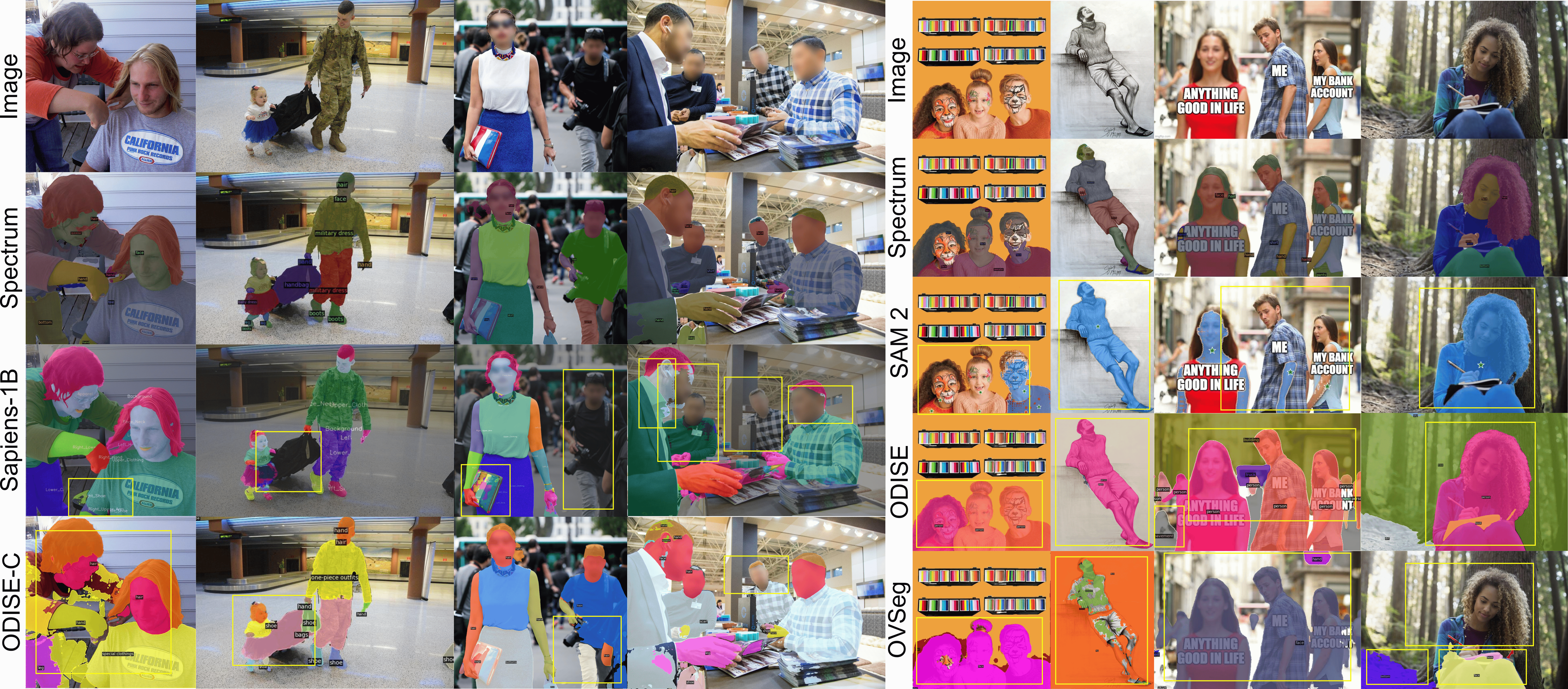}

    \caption{(Left) Multihuman GranD-f: ours surpasses Sapiens-1B and retrained ODISE-C. (Right) Open-vocabulary parsing in the wild: baselines merge all people into one mask, whereas ours preserves distinct clothing and body-part masks.}

    \label{fig:res_glamm}
    \vspace{-10pt}
\end{figure*}

\begin{figure}[t]
  \centering
   \includegraphics[width=0.7\linewidth]{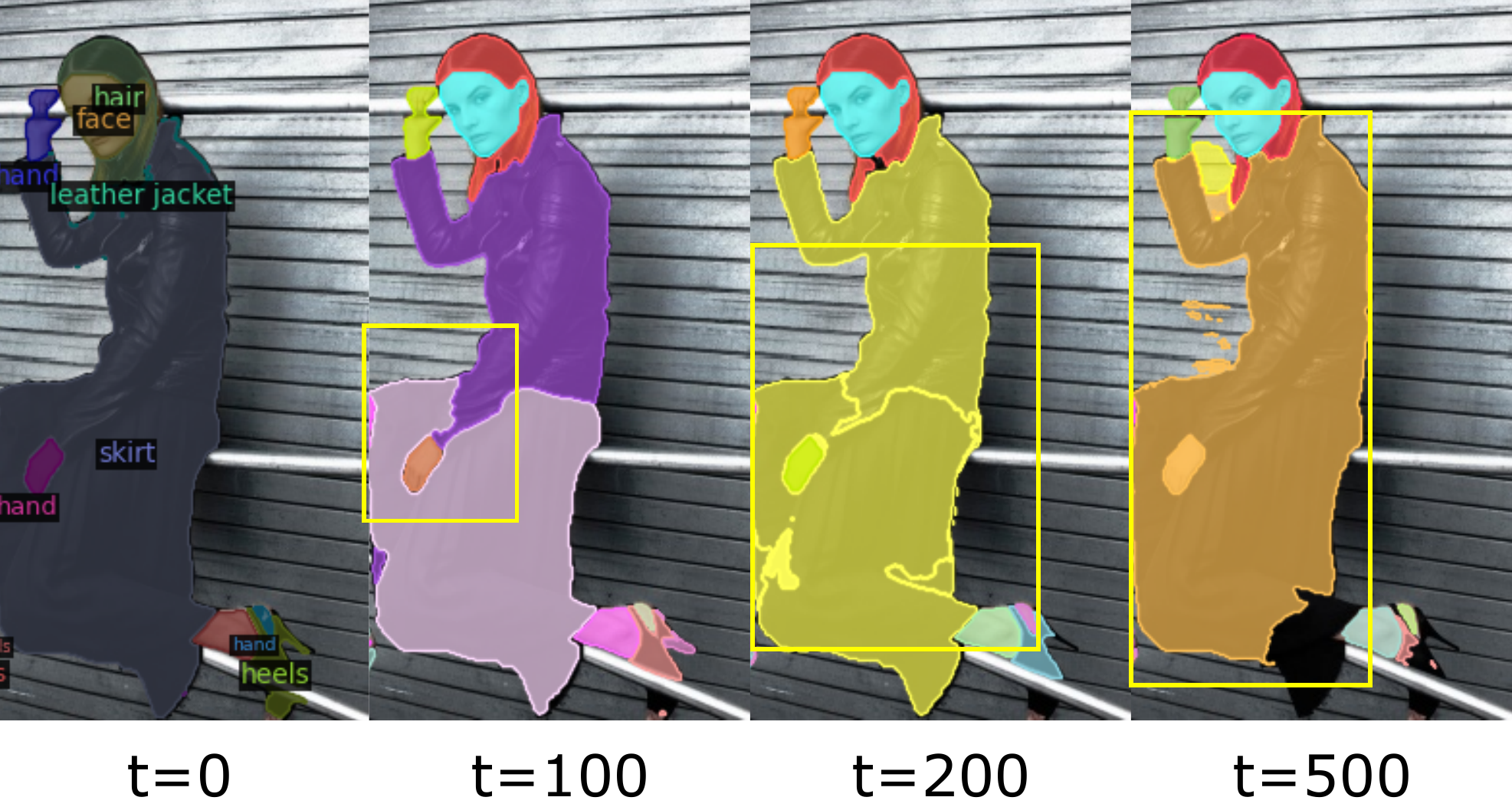}
\caption{Diffusion timesteps ablation. Parsing at $t=0,100,200,500$; $t=0$ performs best.}

   \label{fig:diff_sup}
   \vspace{-10pt}
\end{figure}

%% file: sec/6_experiments.tex
\section{Experiments}
\label{sec:experiments}

\begin{table}[t]
    \centering
    {\scriptsize
    \resizebox{\columnwidth}{!}{%
        \begin{tabular}{l|ccc|ccc}
            \multirow{2}{*}{\textbf{Method}} & \multicolumn{3}{c|}{\textbf{CosmicManHQ -- COP}} & \multicolumn{3}{c}{\textbf{Cross data: GranD-f -- COP}} \\
            & \textbf{mIoU} & \textbf{mAcc} & \textbf{mAP\textsuperscript{SS}} & \textbf{mIoU} & \textbf{mAcc} & \textbf{mAP\textsuperscript{SS}} \\
            \Xhline{.8pt}
            GD$^\dagger$~\cite{li2023grounded} & 14.6 & 18.0 & 11.5 & 7.6 & 9.8 & 6.5 \\
            IFSeg~\cite{yun2023ifseg} & 32.0 & 35.2 & 24.1 & 13.7 & 15.0 & 10.4 \\
            MCLIP$^*$~\cite{ding2023maskclip} & 29.1 & 32.0 & 21.9 & 13.2 & 14.5 & 10.0 \\
            ODISE$^\dagger$~\cite{xu2023odise} & 33.3 & 36.6 & 25.1 & 14.0 & 15.4 & 10.6 \\
            OVSeg$^*$~\cite{liang2023open} & 25.6 & 27.5 & 19.3 & 12.8 & 14.1 & 9.7 \\
            OVAM$^\dagger$~\cite{Marcos-Manchon_2024_CVPR} & 25.6 & 29.5 & 19.3 & 8.5 & 11.8 & 7.9 \\
            SED~\cite{xie2024sed} & \cellcolor{gray!20}44.0 & \cellcolor{gray!20}48.4 & \cellcolor{gray!20}33.1 & \cellcolor{gray!20}14.6 & \cellcolor{gray!20}16.8 & \cellcolor{gray!20}11.0 \\
            SEEM~\cite{zou2023segment} & 29.8 & 36.0 & 21.4 & 11.4 & 12.8 & 9.9 \\
            LISA~\cite{lai2023lisa} & \cellcolor{gray!20}40.5 & \cellcolor{gray!20}44.0 & \cellcolor{gray!20}30.4 & \cellcolor{gray!20}15.2 & \cellcolor{gray!20}17.7 & \cellcolor{gray!20}10.9 \\
            Ours$^\ddagger$ & \cellcolor{pastelblue!60}\textbf{77.5} & \cellcolor{pastelblue!60}\textbf{79.4} & \cellcolor{pastelblue!60}\textbf{58.2} & \cellcolor{pastelblue!60}\textbf{27.1} & \cellcolor{pastelblue!60}\textbf{29.8} & \cellcolor{pastelblue!60}\textbf{21.0} \\
        \end{tabular}%
    }}
    \caption{Open-vocabulary semantic parsing. Legend: $\dagger$ diffusion-based; $*$ CLIP-based; $\ddagger$ uses both methods.}
    \label{tab:open_vocab}
    \vspace{-10pt}
\end{table}

\paragraph{Evaluation settings.} We assess \modelname~under four tasks (see~\cref{fig:eval_setting}):
FPP—Full-Person Parsing (one mask per person),
BHP—Bare-Human Parsing (body part masks),
CCP—COCO Category Parsing (accessories: \textcolor{blue}{\textit{backpack}}, \textcolor{blue}{\textit{umbrella}}, \textcolor{blue}{\textit{shoe}}, \textcolor{blue}{\textit{eye glasses}}, \textcolor{blue}{\textit{handbag}}, \textcolor{blue}{\textit{tie}}, \textcolor{blue}{\textit{suitcase}}), and
COP—Clothing-Only Parsing (one mask per clothing).
To ensure fair comparison, we unify baseline labels by relabelling equivalent categories (e.g., \textcolor{blue}{\textit{pants}}$\!\rightarrow$\textcolor{blue}{\textit{bottoms}}, \textcolor{blue}{\textit{upper/lower arms}}$\!\rightarrow$\textcolor{blue}{\textit{hand}}) and ignore classes a baseline was never trained to predict. \modelname~outputs BHP, CCP, and COP masks, which we merge to form an FPP mask for comparison. All scores use pixel-wise agreement with the GT.

\paragraph{Cross-dataset evaluation.} The model is trained on single-human CosmicManHQ~\cite{li2024cosmicman} and tested on its hold-out split, on GranD-f~\cite{hanoona2023GLaMM} (multi-human, all four settings), and on standard human parsing datasets ATR~\cite{Liang_2015}, LIP~\cite{gong2017lookpersonselfsupervisedstructuresensitive}, PPP~\cite{mottaghi_cvpr14}, and CIHP~\cite{gong2018instancelevelhumanparsinggrouping}. GranD-f poses challenging crowded scenes, heavy occlusion, and incomplete GT masks—resulting in lower absolute scores for all methods. The prompts follow the original annotations (CosmicManHQ: BLIP~\cite{li2022blipbootstrappinglanguageimagepretraining}; GranD-f: Vicuna~\cite{vicuna2023}); the lengthy captions are compressed while retaining every clothing term. Qualitative figures show the prompt (\textcolor{teal}{teal}) and the extended body-parts list (EBP); the full EBP is \textcolor{blue}{\textit{face}}, \textcolor{blue}{\textit{faces}}, \textcolor{blue}{\textit{hand}}, \textcolor{blue}{\textit{hands}}, \textcolor{blue}{\textit{hair}}, \textcolor{blue}{\textit{hairs}}, \textcolor{blue}{\textit{wavy}}, \textcolor{blue}{\textit{ponytail}}, \textcolor{blue}{\textit{bob}}, \textcolor{blue}{\textit{bald}}, \textcolor{blue}{\textit{curly}}, \textcolor{blue}{\textit{afro-hair}}, \textcolor{blue}{\textit{leg}}, \textcolor{blue}{\textit{legs}}, \textcolor{blue}{\textit{back}}, \textcolor{blue}{\textit{chest}}, \textcolor{blue}{\textit{belly}}, \textcolor{blue}{\textit{stomach}}, and \textcolor{blue}{\textit{feet}}.

\begin{table}[t]
    \centering
    {\scriptsize
    \resizebox{\columnwidth}{!}{%
        \begin{tabular}{l|cccc|cccc}
            \multirow{3}{*}{\textbf{Method}} & \multicolumn{4}{c|}{\textbf{CosmicManHQ}} & \multicolumn{4}{c}{\textbf{Cross-dataset: GranD-f}} \\
            & \multicolumn{2}{c}{\textbf{COP}} & \multicolumn{2}{c|}{\textbf{FPP}} & \multicolumn{2}{c}{\textbf{COP}} & \multicolumn{2}{c}{\textbf{FPP}} \\
            & \textbf{mAP\textsuperscript{IS}} & \textbf{AR@100} & \textbf{mAP\textsuperscript{IS}} & \textbf{AR@100} & \textbf{mAP\textsuperscript{IS}} & \textbf{AR@100} & \textbf{mAP\textsuperscript{IS}} & \textbf{AR@100} \\
            \Xhline{.8pt}
            GGN & --- & --- & 20.4 & 33.0 & --- & --- & 11.7 & 26.3 \\
            MCLIP$^*$ & 3.9 & 6.0 & 27.8 & 39.4 & 2.2 & 5.4 & 19.5 & 29.2 \\
            OLN & --- & --- & 23.6 & 30.7 & --- & --- & 11.3 & 24.7 \\
            ODISE$^\dagger$ & 4.2 & 7.7 & 36.5 & 59.5 & 3.6 & 6.9 & 22.9 & 30.1 \\
            Ours$^\dagger$ & \cellcolor{pastelblue!60}\textbf{24.7} & \cellcolor{pastelblue!60}\textbf{30.3} & \cellcolor{pastelblue!60}\textbf{42.4} & \cellcolor{pastelblue!60}\textbf{66.1} & \cellcolor{pastelblue!60}\textbf{17.1} & \cellcolor{pastelblue!60}\textbf{26.2} & \cellcolor{pastelblue!60}\textbf{31.2} & \cellcolor{pastelblue!60}\textbf{39.8} \\
        \end{tabular}%
    }}
    \caption{Open-vocabulary instance parsing.}
    \label{tab:instance_parsing}
    \vspace{-10pt}
\end{table}

\begin{table*}[t]
\centering
{\scriptsize
\setlength{\tabcolsep}{2.5pt}
\renewcommand{\arraystretch}{0.95}
\resizebox{\textwidth}{!}{%
\begin{tabular}{l | c c c c c c c c c c c c c c c c c c}
Model & Face & \multicolumn{2}{c}{$_L\text{Hand}_R$} & Hair & Bags & Special Clothings & Tops & Eyewear & \multicolumn{2}{c}{$_L\text{Leg}_R$} & Hats & Belts & \multicolumn{2}{c}{$_L\text{Shoe}_R$} & One-piece Outfits & Scarf & Bottoms & Avg \\
\Xhline{.8pt}
w/o \textbf{C}& 16.71 & 19.63 & 16.46 & 15.11 & 18.5  & 11.59  & 17.48 & 19.58 & 15.19 & 16.12 & 18.81 & 13.91 & 18.11 & 16.19 & 17.07 & 19.43 & 18.93 & 16.99 \\
w/o $\mathcal{L}_G$    & 52.99 & 62.24 & 52.2  & 47.91 & 58.67 & 36.76  & 55.43 & 62.09 & 48.18 & 51.13 & 59.67 & 44.1  & 57.44 & 51.35 & 54.14 & 61.61 & 60.05 & 53.88 \\
w/o $f$  & 66.25 & 77.82 & 65.26 & 59.9  & 73.35 & 45.95  & 69.3  & 77.63 & 60.23 & 63.93 & 74.59 & 55.13 & 71.82 & 64.2  & 67.68 & 77.02 & 75.07 & 67.36 \\
Ours      & \cellcolor{pastelblue!60}84.47 & \cellcolor{pastelblue!60}91.22 & \cellcolor{pastelblue!60}83.21 & \cellcolor{pastelblue!60}84.37 & \cellcolor{pastelblue!60}93.52 & \cellcolor{pastelblue!60}62.59  & \cellcolor{pastelblue!60}88.36 & \cellcolor{pastelblue!60}94.98 & \cellcolor{pastelblue!60}76.8  & \cellcolor{pastelblue!60}81.51 & \cellcolor{pastelblue!60}95.11 & \cellcolor{pastelblue!60}72.3  & \cellcolor{pastelblue!60}91.57 & \cellcolor{pastelblue!60}81.86 & \cellcolor{pastelblue!60}86.3  & \cellcolor{pastelblue!60}96.21 & \cellcolor{pastelblue!60}95.72 & \cellcolor{pastelblue!60}\textbf{85.89} \\
\end{tabular}%
}}
\caption{Per-class IoU (\%) for \modelname\ and ablations (w/o \textbf{C}, $\mathcal{L}_G$, $f$). L/R subscripts indicate left/right parts.}
\label{tab:per_class_iou}
\vspace{-10pt}
\end{table*}

\begin{table}[t]
    \centering

    \begin{minipage}[t]{0.49\linewidth}
        \centering
        \subcaption{Unseen (U)/ Seen (S) COP}
        \label{tab:unseen_seen}
        {\scriptsize
        \resizebox{\linewidth}{!}{
        \begin{tabular}{l|cc|cc}
            \multirow{2}{*}{\textbf{Method}} & \multicolumn{2}{c|}{\textbf{Cosmic}} & \multicolumn{2}{c}{\textbf{GranD-f}} \\
            & \textbf{U} & \textbf{S} & \textbf{U} & \textbf{S} \\ \Xhline{.8pt}
            ODISE   & 24.6 & 38.5 & 7.7  & 18.1 \\
            ODISE-C & \cellcolor{gray!20}51.2 & \cellcolor{gray!20}69.0 & \cellcolor{gray!20}15.2 & \cellcolor{gray!20}23.3 \\
            Ours    & \cellcolor{pastelblue!60}\textbf{60.8} & \cellcolor{pastelblue!60}\textbf{80.8} & \cellcolor{pastelblue!60}\textbf{18.9} & \cellcolor{pastelblue!60}\textbf{29.9} \\
        \end{tabular}}}
    \end{minipage}\hfill
    \begin{minipage}[t]{0.49\linewidth}
        \centering
        \subcaption{Cross-dataset eval.}
        \label{tab:cross_dataset}
         \vspace{0.9em}
        {\scriptsize
        \resizebox{\linewidth}{!}{
        \begin{tabular}{l|cccc}
            \textbf{Method} & \textbf{LIP} & \textbf{ATR} & \textbf{PPP} & \textbf{CIHP} \\ \Xhline{.8pt}
            SCHP & 59.36 & 82.29 & 71.46 & -- \\
            PGN  & --    & --    & --    & 55.8 \\
            Ours & \cellcolor{pastelblue!60}\textbf{85.6} & \cellcolor{pastelblue!60}\textbf{85.9} & \cellcolor{pastelblue!60}\textbf{82.5} & \cellcolor{pastelblue!60}\textbf{80.3} \\
        \end{tabular}}}
    \end{minipage}

    \caption{Unseen/seen and cross-dataset mIoU evaluation.}
    \label{tab:unseen_seen_cross}
     \vspace{-10pt}
\end{table}

\begin{table}[t]
    \centering
    {\scriptsize
    \setlength{\tabcolsep}{6pt} %
    \renewcommand{\arraystretch}{1.05}
    \begin{tabular}{l|ccc}
        \textbf{Method} & \textbf{FPS} & \textbf{GFLOPS} & \textbf{\#Params} \\ \Xhline{.8pt}
        OVSeg~\cite{liang2023open} & 1.1 & 0.830 & 531M \\
        ODISE~\cite{xu2023odise} & \cellcolor{gray!20}0.6 & \cellcolor{gray!20}0.953 & \cellcolor{gray!20}1522M \\
        MCLIP~\cite{ding2023maskclip} & \cellcolor{pastelblue!60}\textbf{2.5} & \cellcolor{pastelblue!60}\textbf{0.542} & \cellcolor{pastelblue!60}\textbf{367M} \\
        Ours & \cellcolor{gray!20}0.5 & \cellcolor{gray!20}0.976 & \cellcolor{gray!20}1571M \\
    \end{tabular}
    }
    \caption{Model size comparison.}
    \label{tab:model_size_analysis}
     \vspace{-10pt}
\end{table}

\subsection{Closed-Set Segmentors}
We benchmark \modelname~against state of the art human-parsing models that assume a fixed label set (BHP and COP only). Specifically, we compare against (i) the original Sapiens-1B \cite{khirodkar2024sapiensfoundationhumanvision} and CIHP-PGN~\cite{gong2018instancelevelhumanparsinggrouping}; (ii) three SCHP~\cite{li2019selfcorrectionhumanparsing} variants retrained on the standard ATR, LIP, and PPP human parsing datasets; and (iii) two models trained on our CosmicManHQ split—Mask2Former in a closed-set setting (M2F-C\textsubscript{close}) \cite{cheng2021mask2former} and the open-vocabulary segmentor ODISE-C \cite{xu2023odise}. (PPP-SCHP outputs BHP masks only due to label mismatches with COP.)~\Cref{tab:close_set} reports results on the CosmicManHQ test split and on GranD-f. Each COP/BHP entry lists \mbox{(mIoU, mAcc, mAP\textsuperscript{SS}@0.5–0.95)}. \modelname~outperforms all baselines in every metric; Sapiens-1B ranks second, while ODISE-C, though open-vocabulary, still falls short. Cross-dataset results in~\cref{tab:cross_dataset} show that our method achieves higher mIoU than the three SCHP variants on their respective training datasets (LIP, ATR, and PPP) and also surpasses the PGN baseline on CIHP, highlighting its strong generalization ability. Qualitative comparisons in~\cref{fig:cosmic_results} (rows 1–4) reflect this trend: \modelname~delivers complete, semantically correct masks for all visible body parts and novel garments (\textcolor{blue}{\textit{wedding dress}}, \textcolor{blue}{\textit{skirt}}, \textcolor{blue}{\textit{shirt}}), dynamically assigning new colours (e.g., different colours for multiple hands). In comparison, Sapiens-1B and other baselines show fragmented regions (yellow box highlight). See the \supmat for dynamic-color mask assignments and captions for all qualitative results.

\subsection{Open-Vocabulary Segmentors}
Most open-vocabulary models output a single \textcolor{blue}{\textit{person}} mask (FPP) and lack BHP support; we therefore evaluate them only under COP. \textbf{Semantic segmentation.} We benchmark IFSeg~\cite{yun2023ifseg}, MCLIP~\cite{ding2023maskclip}, OVSeg~\cite{liang2023open}, SED~\cite{xie2024sed}, SEEM~\cite{zou2023segment}, LISA~\cite{lai2023lisa}, and the T2I-based GD~\cite{li2023grounded}, OVAM~\cite{Marcos-Manchon_2024_CVPR}, and ODISE~\cite{xu2023odise}. \textbf{Instance segmentation.} We include GGN~\cite{wang2022ggn}, OLN~\cite{kim2021oln}, MCLIP, and ODISE.~\Cref{tab:open_vocab} lists COP scores (mIoU, mAcc, mAP\textsuperscript{SS}@0.5–0.95) on the CosmicManHQ test split and on GranD-f; \modelname~outperforms all semantic baselines, with SED and LISA closest.~\Cref{tab:instance_parsing} reports instance metrics (mAP\textsuperscript{IS}@0.5–0.95, AR@100); \modelname~again leads by a wide margin (GGN, OLN cannot output COP masks). Qualitative comparisons in~\cref{fig:res_glamm} show complex multi-human scenes from GranD-f (left) and in-the-wild images (right). \modelname~yields coherent masks for every garment and body part, whereas parsing baselines (Sapiens-1B, ODISE-C) fragment garments and open-vocabulary segmentors (ODISE, OVSeg) often collapse all people into one mask (highlighted in yellow boxes). We also tested SAM~2~\cite{ravi2024sam2} with point prompts on a small subset; results are shown qualitatively only, as its task setting is not directly comparable.

\subsection{Unseen Clothing}
Because closed-set parsers cannot handle novel classes, we evaluate only open-vocabulary models under COP, separating seen vs. unseen clothing. A BERT-NER filter~\cite{DBLP:journals/corr/abs-1810-04805} removes test-set labels that overlap the training set, leaving categories such as \textcolor{blue}{\textit{maxi}}, \textcolor{blue}{\textit{floral shirt}}, \textcolor{blue}{\textit{jumpsuit}}, \textcolor{blue}{\textit{loafers}}, \textcolor{blue}{\textit{sneakers}}, \textcolor{blue}{\textit{polo shirt}}, \textcolor{blue}{\textit{midi}}, \textcolor{blue}{\textit{camisole}}, \textcolor{blue}{\textit{hoodie}}, \textcolor{blue}{\textit{leather jacket}}, \textcolor{blue}{\textit{slippers}}, \textcolor{blue}{\textit{military dress}}, \textcolor{blue}{\textit{saree}}, \textcolor{blue}{\textit{baby dress}}, \textcolor{blue}{\textit{jersey}}, \textcolor{blue}{\textit{fur coat}}, \textcolor{blue}{\textit{brief}}, \textcolor{blue}{\textit{bandana}}, etc.~\Cref{tab:unseen_seen} reports mIoU on CosmicManHQ and GranD-f for unseen/seen COP splits, comparing \modelname~with two top-performing baselines, ODISE and retrained ODISE-C. \modelname~achieves the highest scores, with ODISE-C second. Qualitative examples in~\cref{fig:teaser,fig:cosmic_results,fig:res_glamm} further show accurate masks on novel clothing, thanks to the repurposed I2Tx features and prompt grounding.

\subsection{Model Size Comparison}
\Cref{tab:model_size_analysis} reports FPS, GFLOPs, and parameters. Although \modelname~is not efficiency-focused, it runs on a single 20GB RTX4000 with compute close to ODISE, while MCLIP has the smallest footprint.

\subsection{Ablation Studies}

We ablate \modelname~on the CosmicManHQ test split and report per-class IoU is in~\cref{tab:per_class_iou}. Additional FPP/CCP metrics, qualitative examples (single/multi-human), prompt variants, limitations, and ethical notes are in the \supmat

\paragraph{Standard \sd~features.} Substituting our I2Tx texture features with vanilla \sd~latent features reduces mIoU to 67.36, showing that texture-aligned latents are substantially more informative for diverse clothing and body-part parsing.

\paragraph{Without context embedding ($\mathbf{C}$ and \textsc{CLS}).} Removing the context embedding drops mIoU from 85.89 to 16.99, as it provides essential conditioning for the I2Tx~model.

\paragraph{Without grounding loss (\(\mathcal{L}_{\text{G}}\)).} Omitting the contrastive prompt-grounding loss drops mIoU to 53.88 and causes clothing to collapse to generic \textcolor{blue}{\textit{Special Clothing}} label (\cref{fig:cosmic_results} R5), highlighting the importance of prompt alignment.

\paragraph{Diffusion timesteps.}
We extract I2Tx features in a single forward pass at $t{=}0$, skipping denoising. Sampling $t \in \{100, 200, 500\}$ from the 1000-step schedule yields fragmented masks and poorer segmentation (\cref{fig:diff_sup}).

\paragraph{Low exposure ablation.}
Our training data (CosmicManHQ split) and ATLAS 3D textures are well lit, so darker scenes create a domain gap. Applying gamma correction ($0<\gamma<1$) to simulate under-exposure drops mIoU to 79.11 at $\gamma=0.75$, 54.20 at $\gamma=0.50$, and 30.90 at $\gamma=0.25$, showing that low light markedly harms performance. 

\paragraph{Prompt dependency.} \modelname~is insensitive to prompt count, mIoU is stable (77.5@\(K_{\text{phrase}}{=}9\) vs.\ 76.8@\(K_{\text{phrase}}{=}5\)), and dropping 30\% of phrases lowers mIoU by only 0.8.

%% file: sec/7_conclusion.tex
\section{Conclusion}
\label{sec:conclusion}
We present \modelname, a unified network for part-level pixel parsing of diverse clothing and body parts, as well as instance-level part grouping. It leverages the internal representation of an I2Tx diffusion model—trained to generate 3D textures from input images—to generate accurate segmentation maps for visible body parts and clothing categories, while ignoring standalone garments and irrelevant objects. Extensive cross-dataset experiments, including single and multi-human scenarios, evaluate body parts, clothing parts, unseen categories, and full-body masks. \modelname~consistently outperforms state-of-the-art human parsers, open-vocabulary segmentors, and instance segmentation baselines in prompt-based mask prediction. In particular, it achieves superior generalization to unseen clothing categories in cross-dataset settings. These findings suggest that repurposing 3D texture features from fine-tuned diffusion models can support more detailed and generalizable human parsing.

%% file: sec/X_suppl.tex
\makeatletter
\renewcommand{\thesection}{\Alph{section}} 
\renewcommand{\thesubsection}{\thesection.\arabic{subsection}}
\renewcommand{\thesubsubsection}{\thesubsection.\arabic{subsubsection}}
\renewcommand{\thefigure}{A\arabic{figure}}
\renewcommand{\thetable}{A\arabic{table}}  
\renewcommand{\theequation}{A\arabic{equation}}

\makeatother

\setcounter{section}{0}
\setcounter{subsection}{0}
\setcounter{subsubsection}{0}
\setcounter{figure}{0}
\setcounter{table}{0}

\setcounter{page}{1}

\noindent The supplementary material provides full FPP and CCP results, details on prompt variants and ensembling, prompts used for the qualitative examples, additional qualitative results (dynamic mask color assignment, single- and multi-human scenes, and challenging cases), and a discussion of limitations and ethical considerations.

\section{Additional Quantitative Results}
\subsection{Baseline Training Details}
\label{sup:baselinemethods}

In~\cref{tab:fpp_ccp_cmhq,tab:fpp_ccp_grandf} we summarise baseline models used in Sec.~Experiments for human parsing, closed-set, and open-vocabulary segmentation. The table lists each model’s training categories (Cats.). Parsing models employ fixed label sets with no support for unseen classes, whereas open-vocabulary models are typically trained on larger COCO Thing (T) and Stuff (S) sets. GD$^\dagger$ and OVAM$^\dagger$ rely on synthetic data; OVAM is trained on only 1,100 COCO-Cap captions, and neither OVAM nor LISA report their training-category counts. Human-parsing baselines were trained on PPP, ATR, CIHP, or LIP; Sapiens was pretrained on Humans-300M and fine-tuned on Humans-2K (H\{300M,2K\}). We retrain M2F-C\(_{\text{close}}\) and ODISE-C\(_{\text{open}}\) on CosmicManHQ (CMHQ). Supervision codes (Supv.): M—GT masks, L—labels, C—captions, G—grounding for predicted masks.

\subsection{FPP and CCP}
\label{sup:fppcpp}

In addition to the metrics in Tabs.~1–6 of the main paper, we report FPP and CCP scores for the human-parsing, closed-set, and open-vocabulary baselines in~\cref{tab:fpp_ccp_cmhq,tab:fpp_ccp_grandf} on CosmicManHQ and GranD-f. Although FPP/CCP masks do not separate body parts or fine-grained clothing—the focus of our work—they provide complementary validation. Each cell lists $(\text{mIoU},\ \text{mAcc},\ \text{mAP}^{\text{SS}}@[0.50{:}0.95])$. Because open-vocabulary methods usually evaluate \textcolor{blue}{\textit{person}} as a single full-body class and many baselines are trained with clothing/accessory labels (e.g., COCO accessories in CCP), we include this comparison. CCP is a subset of COP whose categories are seen during training. As shown, \modelname~surpasses all baselines on every metric; second-best scores are shaded gray. Note that human-parsing models and the retrained closed-set segmentor M2F-C do not output FPP masks, as they predict only body parts, while PPP-SCHP yields only BHP masks due to its limited label set.

\section{Additional Qualitative Results}
\subsection{Dynamic Mask Color Allocation}
\label{sup:viz}

\begin{figure}[t]
  \centering
   \includegraphics[width=\linewidth]{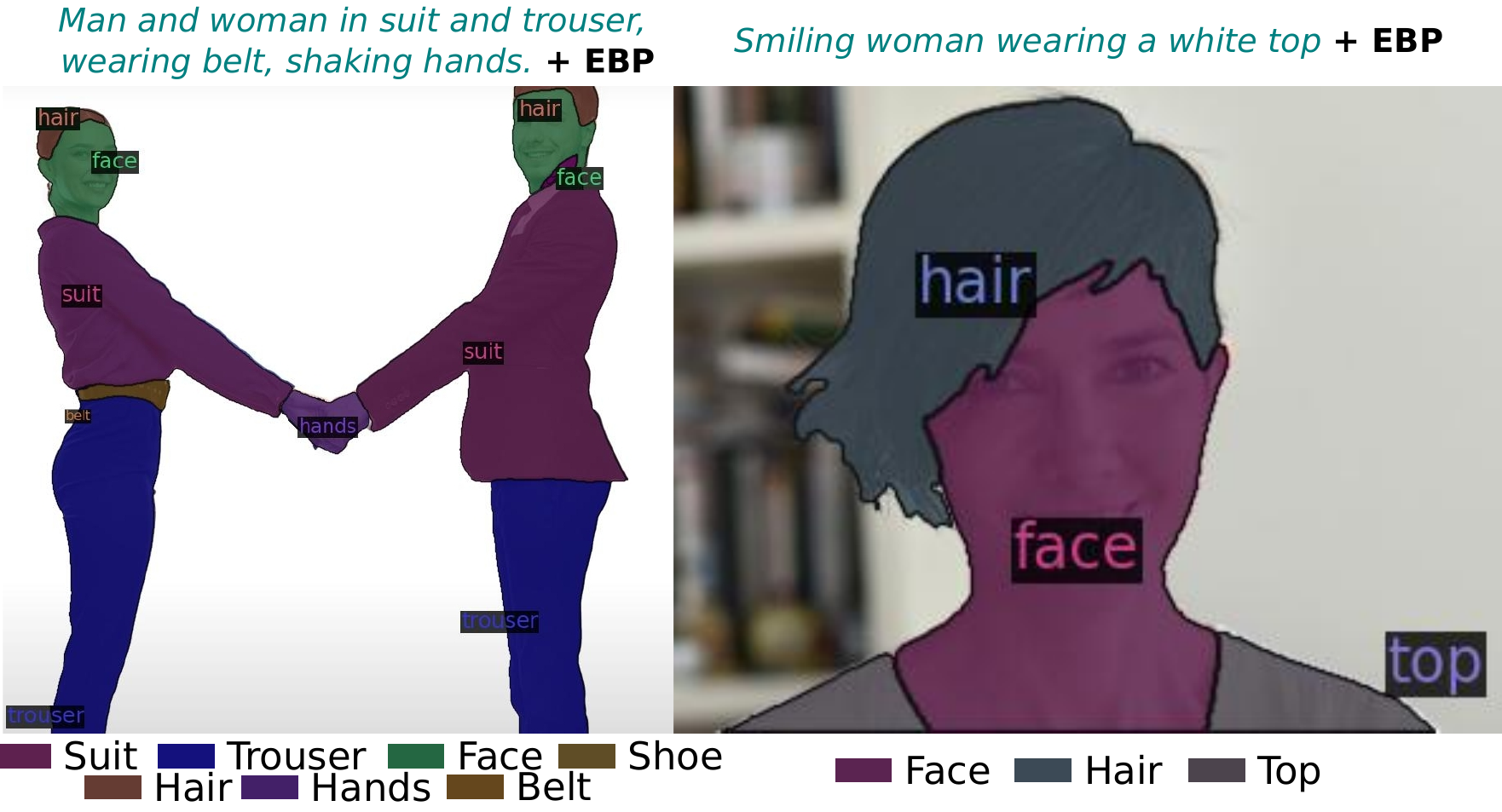}
\caption{Dynamic mask color assignment. Assigned colours appear in the legend below, and the input prompt is shown above.}

   \label{fig:mask_color}
   \vspace{-10pt}
\end{figure}

As noted in Subsec.~Closed-set Segmentors of the main paper, baseline visualisations use their original colour schemes, whereas our method accepts any unseen clothing prompt and dynamically assigns a unique colour to each predicted mask (see~\cref{fig:mask_color}). In the single and multi-human examples, the colours are selected automatically in our code and shown in the legend at the bottom, while the input caption appears at the top.

\subsection{Single and Multi-Human Results}
\label{sup:qual}

In~\cref{fig:multi_fig_sup,fig:single_sup} we provide additional qualitative results on the multi-human GranD-f cross-dataset and the single-human CosmicManHQ test set, respectively. Prompts are shown beside the original images. The examples confirm that \modelname~produces accurate, semantically annotated masks for every visible body part and clothing category, including unseen classes, regardless of the number of humans. It also ignores standalone garments and irrelevant objects, as shown in~\cref{fig:multi_fig_sup} (row~4, col.~1, highlighted in yellow), where a scene with three people and numerous background garments left unsegmented demonstrates the model’s robustness.

\subsection{Challenging Cases}
\label{sup:Challenging}

\begin{figure*}[t]
    \centering
    \captionsetup{type=figure}
 
    \includegraphics[width=1\textwidth]
    {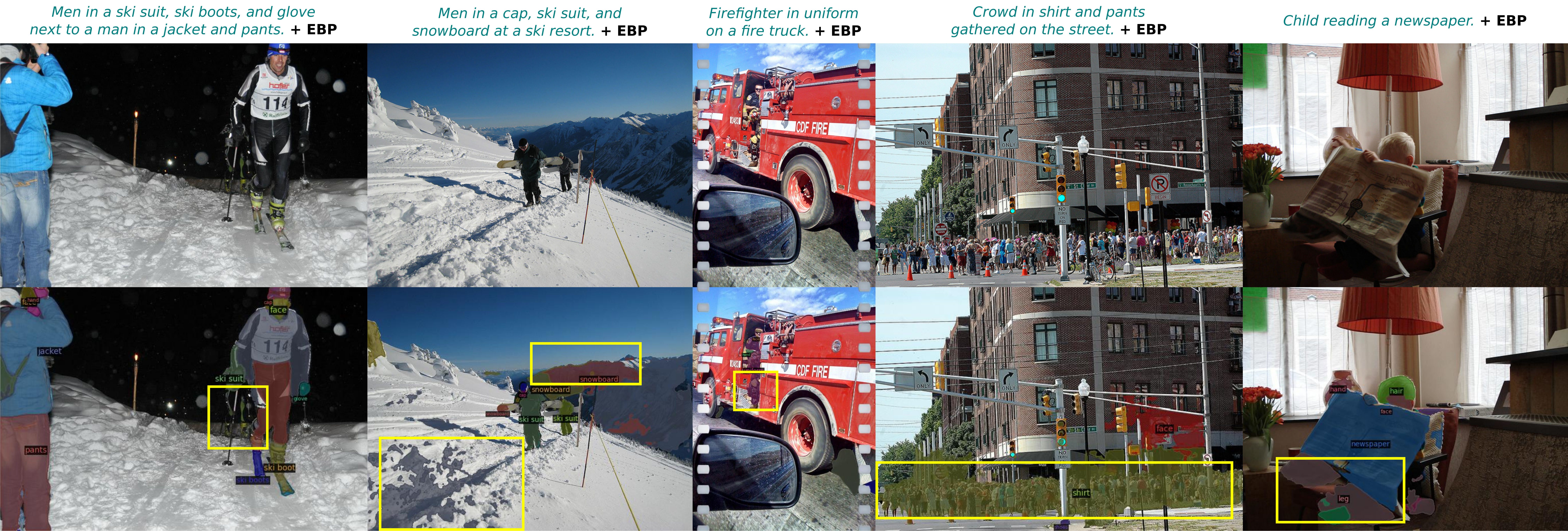}
    \captionof{figure}{Challenging cases. \modelname~encounters difficulties in challenging scenarios involving heavy occlusion, low exposure, complex compositions (e.g., snow-covered mountain regions), and high crowd density.}

    \label{fig:challenge_sup}
      \vspace{-10pt}
\end{figure*}

\begin{figure*}[t]
    \centering
    \captionsetup{type=figure}
 
    \includegraphics[height=.93\textheight,keepaspectratio]
    {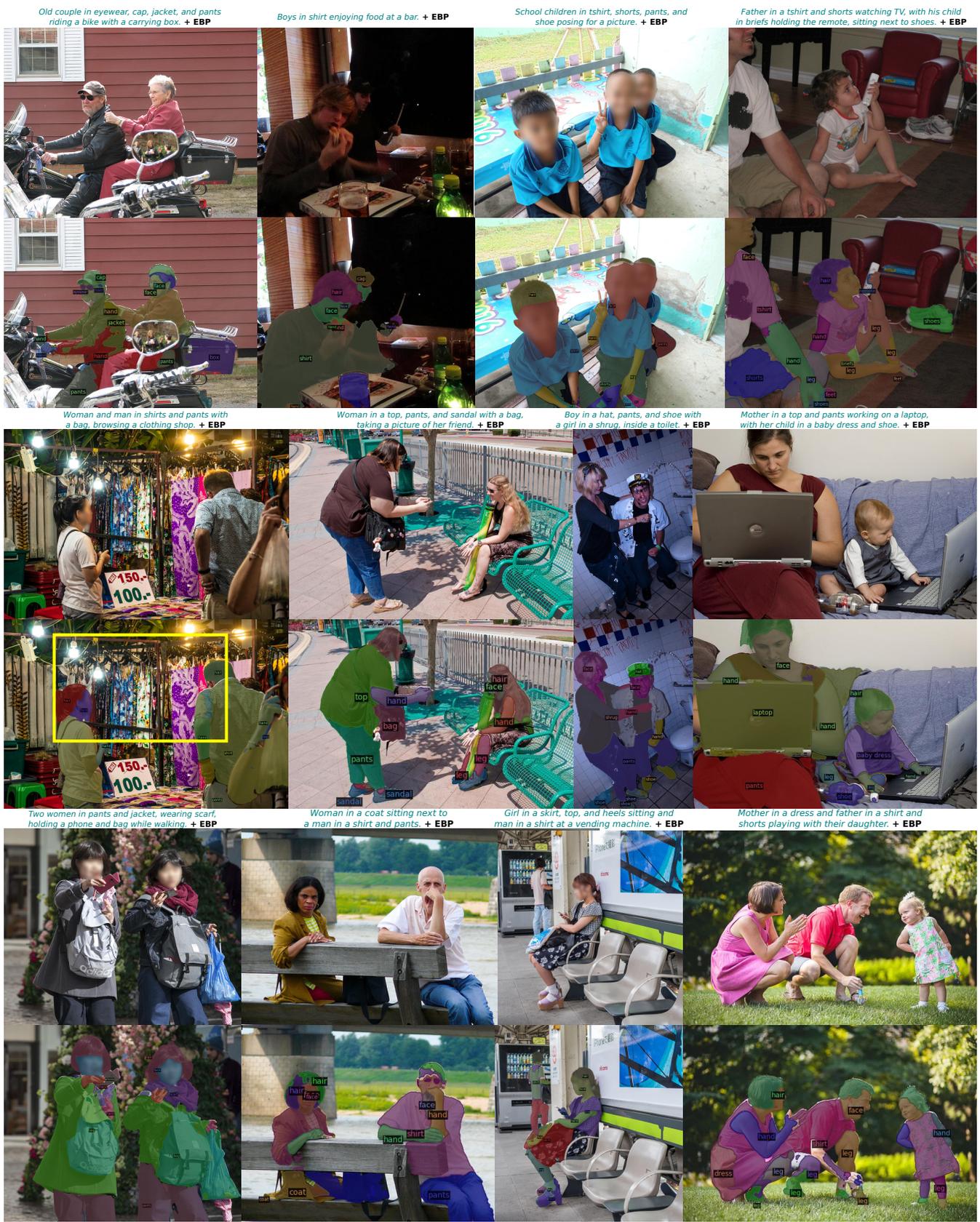}
    \captionof{figure}{Additional qualitative results on GranD-f. In Row~2, Col.~1, \modelname~ignores background garments in a cluttered three-person scene (one person partly occluded) and cleanly parses only the relevant body parts and diverse clothing, demonstrating its robustness.}
    \label{fig:multi_fig_sup}
\end{figure*}

\begin{figure*}[t]
    \centering
    \captionsetup{type=figure}
 
    \includegraphics[width=1\textwidth]
    {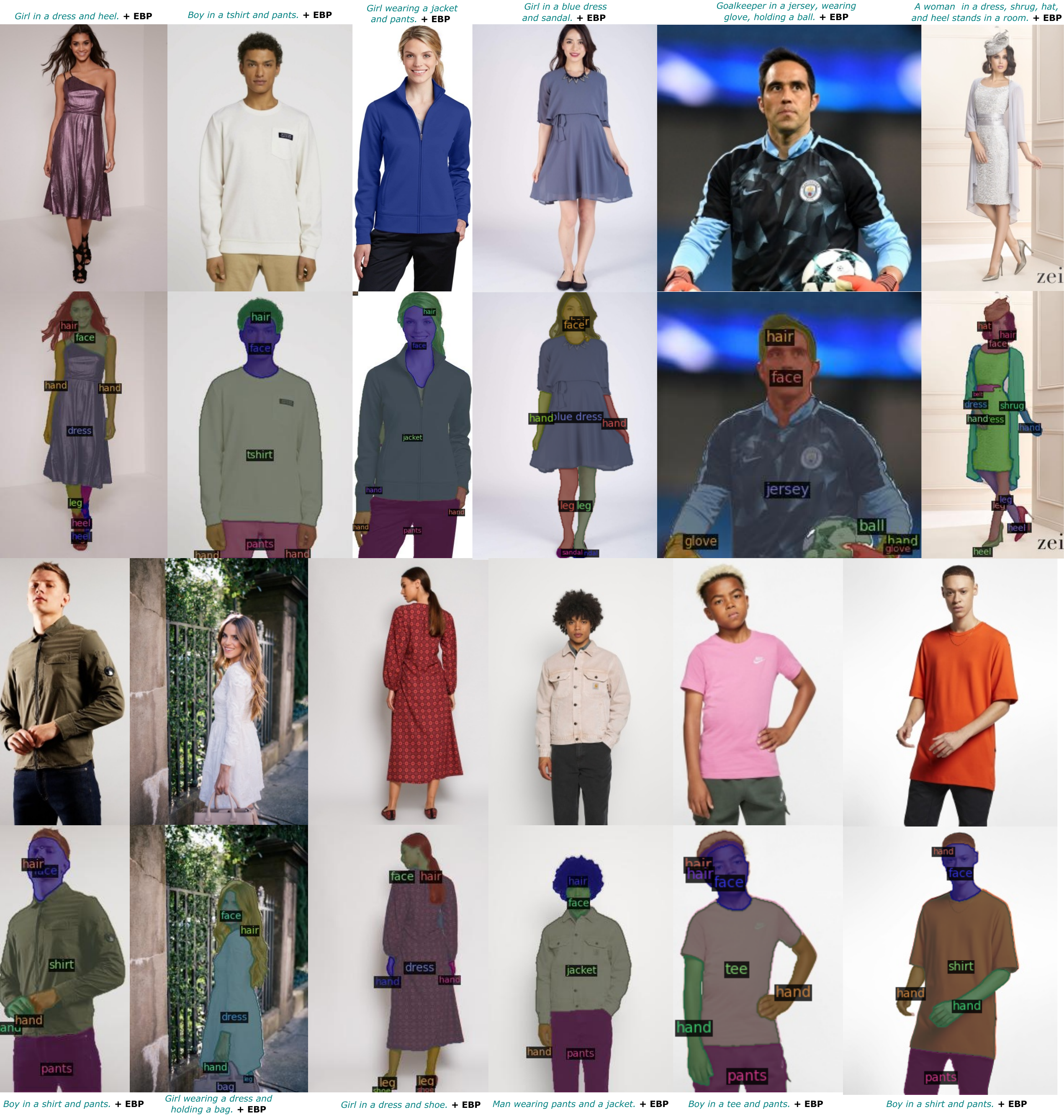}
\captionof{figure}{Additional qualitative results on the CosmicManHQ test set.}
    \label{fig:single_sup}
\end{figure*}

\Cref{fig:challenge_sup} shows challenging cases for \modelname. Columns 1, 3, and 5 involve heavy occlusion of human parts (column 5 also simulates the low-exposure setting from our ablation); column 2 contains a snow-covered background, and column 4 depicts a densely crowded scene.

\begin{table*}[t]
  \centering
  \scriptsize
  \resizebox{\linewidth}{!}{%
    \begin{tabular}{lccccc}
      \multicolumn{1}{c}{\textbf{Method}} &
      \multicolumn{3}{c}{\textbf{Training Details}} &
      \multicolumn{2}{c}{\textbf{CosmicManHQ}} \\
      & \textbf{Cats.} & \textbf{Data} & \textbf{Supv.} & \textbf{FPP} & \textbf{CCP} \\
      \midrule
      PPP-SCHP~\cite{mottaghi_cvpr14}   & 7 & PPP~\cite{mottaghi_cvpr14} & ML & --- & --- \\
      ATR-SCHP~\cite{Liang_2015}   & 18 & ATR~\cite{Liang_2015} & ML & --- & \formatcell{47.3,52.1,35.8} \\
      LIP-SCHP~\cite{gong2017lookpersonselfsupervisedstructuresensitive}   & 20 & LIP~\cite{gong2017lookpersonselfsupervisedstructuresensitive} & ML & --- & \formatcell{44.3,48.1,32.9} \\
      CIHP-PGN~\cite{gong2018instancelevelhumanparsinggrouping} & 20 & CIHP~\cite{gong2018instancelevelhumanparsinggrouping} & ML & --- & \formatcell{45.6,50.4,34.0} \\
      Sapiens-1B~\cite{khirodkar2024sapiensfoundationhumanvision}   & 20,28 & H\{300M,2K\}~\cite{khirodkar2024sapiensfoundationhumanvision} & ML & --- & \cellcolor{gray!20}\formatcell{55.5,59.5,41.0} \\
      M2F-C\textsubscript{close}~\cite{cheng2021mask2former}  & 17 & CMHQ~\cite{li2024cosmicman} & MC & --- & \cellcolor{gray!20}\formatcell{51.5,55.5,33.5} \\
      ODISE-C$^\dagger$\textsubscript{open}~\cite{xu2023odise}   & 17 & CMHQ~\cite{li2024cosmicman} & MCG &
      \cellcolor{gray!20}\formatcell{80.8,81.6,57.9} & \cellcolor{gray!20}\formatcell{69.5,74.1,50.7} \\
      \midrule
      GD$^\dagger$~\cite{li2023grounded}        & 65 & COCO-sim~\cite{li2023grounded} & MLG & \formatcell{40.8,44.9,30.6} & \formatcell{25.0,28.5,16.3 } \\
      IFSeg~\cite{yun2023ifseg}     & 80(T)+91(S) & COCO~\cite{coco} & MC & \formatcell{53.3,58.6,40.0} & \formatcell{45.6,50.1,34.3 } \\
      MCLIP$^*$~\cite{ding2023maskclip}  & 80(T)+53(S) & COCO~\cite{coco} & MLG & \formatcell{48.5,53.4,36.4} & \formatcell{41.5,45.6,31.2} \\
      ODISE$^\dagger$~\cite{xu2023odise}     & 80(T)+53(S) & COCO~\cite{coco} & MCG & \formatcell{55.4,60.9,41.5} & \formatcell{47.4,52.1,35.6 } \\
      OVSeg$^*$~\cite{liang2023open}     & 80(T)+91(S)& COCO~\cite{coco} & MCG & \formatcell{41.7,45.9,31.9} & \formatcell{35.7,39.6,27.7 } \\
      OVAM$^\dagger$~\cite{Marcos-Manchon_2024_CVPR}      &  $\times$  & COCO-cap~\cite{Marcos-Manchon_2024_CVPR} & MCG & \formatcell{49.2,54.1,36.9} & \formatcell{19.1,26.3,16.6} \\
      SED~\cite{xie2024sed}       & 80(T)+91(S)  & COCO~\cite{coco}  & MLG &
      \cellcolor{gray!20}\formatcell{73.4,80.7,55.1 } & \cellcolor{gray!20}\formatcell{62.7,68.9,47.1} \\
      SEEM~\cite{zou2023segment}      & 80  & COCO~\cite{coco} & MLG &
      \cellcolor{gray!20}\formatcell{69.8,76.8,52.3} & \cellcolor{gray!20}\formatcell{39.6,45.5,24.8} \\
      LISA~\cite{lai2023lisa} & $\times$ &
      \makecell[c]{COCO-S~\cite{coco}\\ADE20K~\cite{Zhou_2017_CVPR}\\LVIS-PACO~\cite{ramanathan2023pacopartsattributescommon}} &
      ML &
      \cellcolor{gray!20}\formatcell{71.5,78.0,54.0} & \cellcolor{gray!20}\formatcell{51.2,58.3,36.0} \\
      \midrule
      Ours$^\ddagger$      & 17  & CMHQ~\cite{li2024cosmicman} & MCG &
      \cellcolor{pastelblue!60}\textbf{\formatcell{85.9,88.4,68.8}} &
      \cellcolor{pastelblue!60}\textbf{\formatcell{79.5,82.9,59.3 }} \\
    \end{tabular}
  }
  \caption{\textbf{FPP/CCP on CosmicManHQ.} Top: human parsing and close-set segmentors. Bottom: open-vocabulary segmentors. M2F-C\textsubscript{close} and ODISE-C\textsubscript{open} are retrained on CosmicManHQ. Best/second-best are \colorbox{pastelblue!60}{\textbf{teal}}/\colorbox{gray!20}{gray}. Cells report (mIoU, mAcc, mAP\textsuperscript{SS}@[0.50:0.95]). Diffusion-based methods are denoted by $\dagger$, CLIP-based methods by $*$, and methods using both by $\ddagger$.}

  \label{tab:fpp_ccp_cmhq}
  \vspace{-10pt}
\end{table*}

\begin{table}[t]
  \centering
  \scriptsize
  \resizebox{\linewidth}{!}{%
    \begin{tabular}{lcc}
      \multicolumn{1}{c}{\textbf{Method}} & \multicolumn{2}{c}{\textbf{Cross-dataset: GranD-f}} \\
      & \textbf{FPP} & \textbf{CCP} \\
      \midrule
      PPP-SCHP~\cite{mottaghi_cvpr14}   & --- & --- \\
      ATR-SCHP~\cite{Liang_2015}   & --- & \formatcell{21.2,23.3,16.0} \\
      LIP-SCHP~\cite{gong2017lookpersonselfsupervisedstructuresensitive}   & --- & \formatcell{20.1,22.0,15.1} \\
      CIHP-PGN~\cite{gong2018instancelevelhumanparsinggrouping} & --- & \formatcell{20.7,22.7,15.6} \\
      Sapiens-1B~\cite{khirodkar2024sapiensfoundationhumanvision}   & --- & \cellcolor{gray!20}\formatcell{26.3,29.3,19.3} \\
      M2F-C\textsubscript{close}~\cite{cheng2021mask2former}  & --- & \cellcolor{gray!20}\formatcell{24.5,26.5,17.5} \\
      ODISE-C$^\dagger$\textsubscript{open}~\cite{xu2023odise}   &
      \cellcolor{gray!20}\formatcell{24.3,20.1,15.3 } & \cellcolor{gray!20}\formatcell{22.1,18.2,13.1 } \\
      \midrule
      GD$^\dagger$~\cite{li2023grounded}        & \formatcell{20.7,22.8,15.5 } & \formatcell{12.9,14.7,8.5} \\
      IFSeg~\cite{yun2023ifseg}     & \formatcell{22.6,24.9,17.0} & \formatcell{19.5,21.4,14.7} \\
      MCLIP$^*$~\cite{ding2023maskclip}  & \formatcell{21.6,23.8,16.2} & \formatcell{18.7,20.5,14.1} \\
      ODISE$^\dagger$~\cite{xu2023odise}     & \formatcell{23.1,25.4,17.3} & \formatcell{19.9,21.9,15.0} \\
      OVSeg$^*$~\cite{liang2023open}     & \formatcell{21.1,23.2,15.8} & \formatcell{18.2,20.0,13.7} \\
      OVAM$^\dagger$~\cite{Marcos-Manchon_2024_CVPR}      & \formatcell{22.1,24.3,16.6} & \formatcell{13.5,18.0,11.4} \\
      SED~\cite{xie2024sed}       &
      \cellcolor{gray!20}\formatcell{24.1,26.5,18.1 } & \cellcolor{gray!20}\formatcell{20.8,22.9,15.7} \\
      SEEM~\cite{zou2023segment}      &
      \cellcolor{gray!20}\formatcell{23.6,26.0,17.7} & \cellcolor{gray!20}\formatcell{17.4,19.4,14.4} \\
      LISA~\cite{lai2023lisa} &
      \cellcolor{gray!20}\formatcell{24.0,26.4,18.0} & \cellcolor{gray!20}\formatcell{19.0,21.0,15.2} \\
      \midrule
      Ours$^\ddagger$      &
      \cellcolor{pastelblue!60}\textbf{\formatcell{31.0,34.9,23.8}} &
      \cellcolor{pastelblue!60}\textbf{\formatcell{27.5,30.0,21.7}} \\
    \end{tabular}
  }
  \caption{Cross-dataset FPP/CCP evaluation on GranD-f. Same setup as Tab.~\ref{tab:fpp_ccp_cmhq}.}
  \label{tab:fpp_ccp_grandf}
  \vspace{-10pt}
\end{table}

\section{Prompts}
\label{sup:embseble}

\subsection{Prompt Ensembling}

During inference, the model takes an image $x$ and a natural-language description in one of several forms—open-vocabulary labels (e.g., \textcolor{blue}{\textit{sarees}}, \textcolor{blue}{\textit{military uniform}}), a caption (e.g., “\textcolor{blue}{\textit{Man in military dress and boots holding a handbag, walking with a baby in a dress}}”), or ensemble labels~\cite{ghiasi2022scalingopenvocabularyimagesegmentation} (e.g., \textcolor{blue}{\textit{jumpsuit}}, \textcolor{blue}{\textit{wedding dress}} as novel terms under \textcolor{blue}{\textit{one-piece outfit}})—together with Extended Body Part (EBP) labels to handle prompts that omit body parts. Ensemble labels comprise lists of sub-categories, synonyms, and plurals of the original training classes. We provide ensembling examples for CosmicMan-HQ mask categories, including \textcolor{blue}{\textit{bottom}}, \textcolor{blue}{\textit{one-piece outfit}}, \textcolor{blue}{\textit{hat}}, and \textcolor{blue}{\textit{special clothing}}.

\begin{itemize}
    \item \textcolor{blue}{\textit{bottom}} $\rightarrow$ \textcolor{blue}{\textit{pants}}, \textcolor{blue}{\textit{trousers}}, \textcolor{blue}{\textit{jeans}}, \textcolor{blue}{\textit{shorts}}, \textcolor{blue}{\textit{sweatpants}}, \textcolor{blue}{\textit{bikini bottom}}, \textcolor{blue}{\textit{skirt}}, \textcolor{blue}{\textit{leggings}}, \textcolor{blue}{\textit{cargo pants}}, \textcolor{blue}{\textit{mini}}, \textcolor{blue}{\textit{short}}, \textcolor{blue}{\textit{knee-high}}, \textcolor{blue}{\textit{high-low}}, \textcolor{blue}{\textit{straight}}, \textcolor{blue}{\textit{flare}}, \textcolor{blue}{\textit{close-fitting}}, \textcolor{blue}{\textit{tapered}}, \textcolor{blue}{\textit{wide-leg}}
    \item \textcolor{blue}{\textit{one-piece outfit}} $\rightarrow$ \textcolor{blue}{\textit{dress}}, \textcolor{blue}{\textit{dresses}},  \textcolor{blue}{\textit{one-piece swimsuit}}, \textcolor{blue}{\textit{bathrobe}}, \textcolor{blue}{\textit{romper}}, \textcolor{blue}{\textit{bodysuit}}, \textcolor{blue}{\textit{wedding dress}}, \textcolor{blue}{\textit{jumpsuit}}
    \item \textcolor{blue}{\textit{hat}} $\rightarrow$ \textcolor{blue}{\textit{cap}}, \textcolor{blue}{\textit{caps}}, \textcolor{blue}{\textit{beret}}, \textcolor{blue}{\textit{sun hat}}, \textcolor{blue}{\textit{beanie hat}}, \textcolor{blue}{\textit{bucket hat}}, \textcolor{blue}{\textit{helmet}}, \textcolor{blue}{\textit{baseball cap}}, \textcolor{blue}{\textit{cowboy hat}}
    \item \textcolor{blue}{\textit{special clothing}} $\rightarrow$ \textcolor{blue}{\textit{costume}}, \textcolor{blue}{\textit{costumes}}, \textcolor{blue}{\textit{hanfu}}, \textcolor{blue}{\textit{taekwondo uniform}},  \textcolor{blue}{\textit{graduation gown}}, \textcolor{blue}{\textit{kimono}}, \textcolor{blue}{\textit{police uniform}}, \textcolor{blue}{\textit{cheongsam}}, \textcolor{blue}{\textit{cosplay}}, \textcolor{blue}{\textit{traditional costumes}}, \textcolor{blue}{\textit{fire suit}}, \textcolor{blue}{\textit{hazmat suit}}, \textcolor{blue}{\textit{judicial robe}}
\end{itemize}

\subsection{Input Prompts}

All prompts used in the main paper for qualitative results are listed below. They correspond to Figs.~1, 5, and 6, with panels numbered left to right as 1.1--1.8 (Fig.~1), 5.1--5.5 (Fig.~5), and 6.1--6.8 (Fig.~6). Each prompt is used together with the Extended Body Parts list described in Sec.~Experiments (Cross-dataset evaluation) of the main paper.

1.1 \textcolor{blue}{\textit{Woman in shirt, pants, shoe, standing in an S shape.}}
1.2 \textcolor{blue}{\textit{Woman in shirt, pants, belt, shoes, making a P shape.}}
1.3 \textcolor{blue}{\textit{Woman in shirt, trouser, belt, shoe, holding a file, and man in suit, trouser, shoe, both making an E shape.}}
1.4 \textcolor{blue}{\textit{Boy in tee, pants, and socks, sitting on a stool.}}
1.5 \textcolor{blue}{\textit{Man in jacket, trouser, and shoes, back view, raising his hands.}}
1.6 \textcolor{blue}{\textit{Girl in tshirt, jeans, standing in an R shape.}}
1.7 \textcolor{blue}{\textit{Smiling lady in a saree, raising both hand.}}
1.8 \textcolor{blue}{\textit{Man and woman in suit and trouser, wearing belt, shaing hands.}}

5.1 \textcolor{blue}{\textit{Woman in a wedding dress, wearing a flower hairband and sandal.}}
5.2 \textcolor{blue}{\textit{Girl in boots, skirt, and a sweater.}}
5.3 \textcolor{blue}{\textit{Boy in shirt, shorts, and a handbag across his chest.}}
5.4 \textcolor{blue}{\textit{Girl in a dress applying makeup.}}
5.5 \textcolor{blue}{\textit{Girl in a crop top, dressed for a themed event.}}

6.1 \textcolor{blue}{\textit{Lady in eyewear cutting a man's hair, both in tees and bottom.}}
6.2 \textcolor{blue}{\textit{Father in a military dress and boots walks with his baby in a baby dress.}}
6.3 \textcolor{blue}{\textit{Woman in a skirt and a top with a bag alongside a man in pants in a croweded area.}}
6.4 \textcolor{blue}{\textit{Men in shirt gathered in a meeting.}}
6.5 \textcolor{blue}{\textit{Children in sweater with face painting.}}
6.6 \textcolor{blue}{\textit{A sketch of a man in a sweater, shorts, and slippers.}}
6.7 \textcolor{blue}{\textit{Funny meme of a woman and man in shirt and pants.}}
6.8 \textcolor{blue}{\textit{Girl in nature, wearing a top and bottom, writing in a book with a pen.}}

\section{Limitations}
\label{sup:limits}

Although \modelname~advances diverse clothing and body-part parsing in multi-human scenes, several limitations remain:

\paragraph{Ground-truth masks.} Existing datasets (ATR, PPP, LIP, CIHP, CosmicManHQ, GranD-f) offer a baseline, but large-scale collections with consistent high-quality part masks for every person—foreground and background—in real-world scenes are still needed.

\paragraph{Crowded scenes.} \modelname~handles multi-person scenes, but very dense crowds remain challenging (fig. A1).

\paragraph{Evolving fashion.} Fashion changes rapidly; continual-learning updates will be required to recognise new and shifting clothing styles.

\section{Ethical Considerations}
\label{sup:ethic}

We build on publicly available models (Stable Diffusion, OpenCLIP, TexDreamer, BERT, and CLIP variants) and datasets (CosmicManHQ, GranD-f, ATLAS, ATR, LIP, PPP, CIHP, LAION). Although our method is not inherently designed for discriminatory use, any biases present in these public resources can propagate to our trained model.

